\PassOptionsToPackage{table,xcdraw}{xcolor}

\documentclass[10pt]{article} 
\usepackage[accepted]{rlc}

\usepackage{amssymb}            
\usepackage{mathtools}          
\usepackage{mathrsfs}           

\mathtoolsset{showonlyrefs}    
\usepackage[pdftex]{graphicx}  
        
\usepackage{subcaption}        
\usepackage[space]{grffile}    
\usepackage{url}               

\usepackage[acronym]{glossaries}
\usepackage{multirow}
\usepackage{pdfpages}
\usepackage{amsmath}
\usepackage{paralist}
\usepackage{amsthm}
\usepackage{siunitx}
\usepackage{placeins}
\usepackage{float}
\usepackage{wrapfig}
\usepackage{siunitx}
\usepackage{subcaption}

\theoremstyle{definition}
\newtheorem*{definition*}{Definition}

\title{Combining Automated Optimisation of \\ Hyperparameters and Reward Shape}

\author{Julian Dierkes\\
dierkes@aim.rwth-aachen.de\\
Chair for AI Methodology\\
RWTH Aachen University
\And
Emma Cramer\\
emma.cramer@dsme.rwth-aachen.de\\
Institute for Data Science in Mechanical Engineering\\
RWTH Aachen University
\And
Holger H. Hoos\\
hh@aim.rwth-aachen.de\\
Chair for AI Methodology\\
RWTH Aachen University\\
LIACS, Universiteit Leiden
\And
\hspace{20.5mm}Sebastian Trimpe\\
\hspace{20.5mm}trimpe@dsme.rwth-aachen.de\\
\hspace{20.5mm}Institute for Data Science in Mechanical Engineering\\
\hspace{20.5mm}RWTH Aachen University}

\begin{document}

\maketitle

\begin{abstract}

There has been significant progress in deep reinforcement learning (RL) in recent years. 
Nevertheless, finding suitable hyperparameter configurations and reward functions remains challenging even for experts, and performance heavily relies on these design choices. 
Also, most RL research is conducted on known benchmarks where knowledge about these choices already exists. 
However, novel practical applications often pose complex tasks for which no prior knowledge about good hyperparameters and reward functions is available, thus necessitating their derivation from scratch.
Prior work has examined automatically tuning either hyperparameters or reward functions individually. 
We demonstrate empirically that an RL algorithm's hyperparameter configurations and reward function are often mutually dependent, meaning neither can be fully optimised without appropriate values for the other.
We then propose a methodology for the combined optimisation of hyperparameters and the reward function. 
Furthermore, we include a variance penalty as an optimisation objective to improve the stability of learned policies.
We conducted extensive experiments using Proximal Policy Optimisation and Soft Actor-Critic on four environments.
Our results show that combined optimisation significantly improves over baseline performance in half of the environments and achieves competitive performance in the others, with only a minor increase in computational costs.
This suggests that combined optimisation should be best practice.

\end{abstract}

\section{Introduction}
\label{sec:submission}

Deep reinforcement learning (RL) has successfully been applied to various domains, including \citet{silver17, akkaya19, kaufmann23, bi2023}. 
Despite successes in these and other challenging applications, configuring RL algorithms remains difficult.
This is due to the algorithms typically having several hyperparameter and reward configurations, critically determining learning speed and the general outcome of the training process.
For each task, there usually is a final objective one wants to achieve. 
Defining the RL rewards in terms of this objective is typically insufficient; instead, augmenting the reward with additional intermediate rewards, sub-goals, and constraints is necessary 
for effective training.
This augmentation of a reward signal is referred to as reward shaping, and performance and learning speed can crucially depend on it \citep{ng99}. 
Next, RL algorithms require the optimisation of hyperparameters, such as learning rate or discount factor.
Effective hyperparameter tuning requires an effective reward signal, and effective reward shaping depends on good hyperparameter configurations. 
This circular dependency becomes particularly relevant when applying RL to novel environments beyond commonly used benchmarks, for which neither effective reward shapes nor good hyperparameter settings are available.

In the area of Automatic RL (AutoRL) \citep{parker22}, different data-driven approaches have been developed in recent years to automatically approach hyperparameter optimisation \citep{parker20, falkner18} and reward shaping \citep{wang22, zheng18}.
However, these methods approach each problem individually without considering their interdependency. 
Therefore, they require the availability of high-performing configurations of the non-optimised component.
To the best of our knowledge, ours is the first study to thoroughly investigate the effectiveness and broader applicability of jointly optimising hyperparameters and reward shape by using multiple and different environments and systematically evaluating the benefit thus obtained.

We examine the combined optimisation of hyperparameters and reward shape using two state-of-the-art RL algorithms: Proximal Policy Optimisation (PPO) \citep{schulman17} and Soft Actor-Critic (SAC) \citep{haarnoja18}.
We performed experiments on Gymnasium LunarLander \citep{gym23}, Google Brax Ant and Humanoid \citep{brax21}, and Robosuite Wipe \citep{robosuite20}.
The Wipe environment is a robot task representing contact-rich interactions inspired by modern production tasks, which has not been well studied in the literature yet.
We compare the combined optimisation results against baselines from the literature and, in particular, to individual optimisation of only hyperparameters and reward shape.
We employ the state-of-the-art black-box hyperparameter optimisation algorithm DEHB \citep{awad21} for our experiments, which recently showed to outperform other optimisation methods in RL \citep{eimer23}.

Our key contributions can be summarised as follows:
\begin{compactenum}
    \item We illustrate the advantage of joint optimisation by showing complex dependencies between hyperparameters and reward signals in the LunarLander environment.
    We use an existing hyperparameter optimisation framework and extend it with additional hyperparameters that control reward shaping.
    We show that combined optimisation can match the performance of individual optimisation with the same compute budget despite the larger search space; 
    furthermore, we show that it can yield significant improvement in challenging environments, such as Humanoid and Wipe. 
    \item We demonstrate that including a variance penalty for multi-objective optimisation can obtain hyperparameter settings and reward shapes that substantially improve performance variance of a trained policy while achieving similar or better expected performance.
\end{compactenum}

\section{Background}

We begin with some background on RL, define the optimisation of hyperparameters and reward shape, and present the selected algorithm applicable to these optimisation problems.

\subsection{Reinforcement Learning and Reward-Shaping}
\label{sec:rl_background}

In RL, an agent learns to optimise a \emph{task objective} through interaction with an environment \citep{sutton2018}. 
The environment is represented as a discounted Markov Decision Process (MDP) $\mathcal{M} := (\mathcal{S}, \mathcal{A}, p, r, \rho_0, \gamma)$, with state space $\mathcal{S}$, action space $\mathcal{A}$, an unknown transition probability distribution $p: \mathcal{S} \times \mathcal{A} \times \mathcal{S} \rightarrow \mathbb{R}$, reward function $r: \mathcal{S} \times \mathcal{A} \rightarrow \mathbb{R}$, distribution of the initial state $\rho: \mathcal{S} \rightarrow \mathbb{R}$, and  discount rate $\gamma \in (0, 1)$. 
A policy $\pi: \mathcal{S} \times \mathcal{A} \rightarrow \mathbb{R}$ selects an action with a certain probability for a given state. 
The agent interacts with the MDP to collect episodes $\tau = (s_0, a_0, r_1, s_1, \dots, s_T)$, i.e., sequences of states, actions, and rewards over time steps $t=0, \dots, T$. 

In applications, RL algorithms are very sensitive to the rewards in a given MDP to infer policies achieving the desired objective. 
To ease the policy search, reward shaping is the practice of designing a reward function $\tilde{r}^{\alpha, w} := \alpha \cdot (r + f^{w})$ based on the original reward $r$ of $\mathcal{M}$, where the reward shaping function $f^{w}: \mathcal{S} \times \mathcal{A} \rightarrow \mathbb{R}$ denotes the change in reward \citep{ng99} parameterised by reward weights $w \in \mathbb{R}^n$ and scaled by $\alpha \in \mathbb{R}^+$.
The shaped reward essentially yields the modified MDP $\mathcal{M}_{\alpha, w} := (\mathcal{S}, \mathcal{A}, p, \tilde{r}^{\alpha, w}, \rho_0, \gamma)$.
The function $f^{w}$ is commonly designed by identifying key terms or events that should be rewarded or penalised and combining these as a weighted sum. 

To obtain policies, there are now two hierarchical objectives for measuring performance.
The outer task objective measures success in terms of the overall goal one wants to solve, and the inner objective in terms of maximising the collected shaped rewards when interacting with $\mathcal{M}_{\alpha, w}$.
We formalise the overall task objective as $o_\text{goal}$, measuring the success of a task in the trajectory $\tau$ by assigning it a score $o_\text{goal}(\tau) \in \mathbb{R}$.
Examples of such goals are achieving a certain objective or minimising time to perform a task. 
In addition to this outer task objective, we have the inner objective of maximising the expected return of the shaped rewards given by $J(\pi) = \mathbb{E}_{\tau \sim \pi} [\,\sum_{t=1}^{T} \gamma^t \cdot \tilde{r}_t\,]$.
The common approach of RL is to maximise performance with regard to the task's objective $o_\text{goal}$ by finding the policy $\pi$ that maximizes the expected return $J(\pi)$.
This typically involves tuning the parameters $\alpha$ and $w$ of the shaped reward to obtain reward signals that facilitate finding policies in RL training that perform well with regard to $o_\text{goal}$.
The task objective $o_\text{goal}$ is not used during RL training and only measures success for a full trajectory $\tau$. 
This allows measuring success much sparser than the shaped reward.
Such sparse task objectives are commonly straightforward to define.

\subsection{Combined Hyperparameter and Reward Shaping Optimisation}
\label{back:combined}
\begin{wrapfigure}[23]{R}{0.3\textwidth}
    \center
    \includegraphics[width=0.25\textwidth]{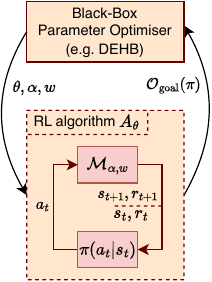}
    \caption{
        Illustration of the two-level optimisation process. Outer loop: hyper- and reward parameter optimisation; inner loop: RL training.
        In each iteration, the parameter optimiser chooses parameters and receives their performance measured by $\mathcal{O}_{\text{goal}}(\pi)$.
    }
    \label{fig:optimisation_loop}
\end{wrapfigure}
In practical RL applications, both the algorithm's hyperparameters and the environment's reward shape require tuning.
For an environment $\mathcal{M}_{\alpha, w}$ with task objective $o_\text{goal}$, we can approach the refinement of hyper- and reward parameters as a two-level optimisation process. 
In the outer loop, an optimisation algorithm selects hyper- and reward parameters for the algorithm and environment. 
In the inner loop, these parameters are used for RL training, yielding a policy $\pi$. 
This policy is then assessed against a task performance metric $\mathcal{O}_\text{goal}$ based on $o_\text{goal}$, and its score is returned to the optimisation algorithm to determine the next parameter configuration. 

To evaluate a policy's performance with regard to $o_\text{goal}$, different metrics can be used. 
The single-objective performance metric $\mathcal{O}^{so}_\text{goal}(\pi):= \mathbb{E}_{\tau \sim \pi}[o_\text{goal}(\tau)]$ is exclusively concerned with optimising the average task score.
The multi-objective metric $\mathcal{O}^{mo}_\text{goal}(\pi):= \mathbb{E}_{\tau \sim \pi}[o_\text{goal}(\tau)] - \sigma_{\tau \sim \pi}[o_\text{goal}(\tau)]$ includes an additional variance-based penalty, as described by \citet{garcia15}, preferring policies with low-performance variance and therefore consistent outcomes.
Figure \ref{fig:optimisation_loop} illustrates the two-level optimisation process. 
To formally introduce the optimisation problems, we adopt the definition of algorithm configuration by \citet{eggensperger19} and adapt it to our RL context. 
Consequently, our focus is on optimising the hyperparameters of the RL algorithm, represented by $\theta$, as well as the reward shaping, represented by $\alpha$ and $w$.
\begin{definition*}
    \label{def:ac}
    Consider an environment $\mathcal{M}_{\alpha, w} := (\mathcal{S}, \mathcal{A}, p, \tilde{r}^{\alpha, w}, \rho_0, \gamma)$ with \emph{reward parameters} consisting of \emph{reward scaling} $\alpha \in A$ and \emph{reward weights} $w \in W$. 
    Further, given an RL algorithm $A_{\theta}(\mathcal{M}_{\alpha, w})$ parametrised by \emph{hyperparameters} $\theta \in \Theta$.
    This algorithm interacts with the environment $\mathcal{M}_{\alpha, w}$ and returns a policy $\pi$.
    For \emph{performance metric} $\mathcal{O}_\text{goal}(\pi)$, we define the following optimisation problems:

    \textbf{Hyperparameter optimisation:} For fixed reward parameters $\hat{\alpha}$ and $\hat{w}$, find $\theta^* \in \Theta$, s.t.
    \begin{equation}
        \theta^* \in \underset{{\theta \in \Theta}}{\text{arg max}}\ \mathcal{O}_\text{goal}(A_{\theta}(\mathcal{M}_{\hat{\alpha},\hat{w}})).
    \end{equation}
    \textbf{Reward parameter optimisation:} For fixed hyperparameters $\hat{\theta}$, find $(\alpha^*, w^*) \in A \times W$, s.t.
    \begin{equation}
        (\alpha^*, w^*) \in \underset{(\alpha, w) \in A \times W}{\text{arg max}}\ \mathcal{O}_\text{goal}(A_{\hat{\theta}}(\mathcal{M}_{\alpha, w})).
    \end{equation} 
    \textbf{Combined optimisation:} Find $(\theta^*, w^*, \alpha^*) \in \Theta \times W \times A$, s.t.
    \begin{equation}
        (\theta^*, w^*, \alpha^*) \in \underset{(\theta, \alpha, w) \in \Theta \times A \times W}{\text{arg max}}\ \mathcal{O}_\text{goal}(A_{\theta}(\mathcal{M}_{\alpha, w})).
    \end{equation}
\end{definition*}

\subsection{DEHB}

Among the many optimisation methods for RL, DEHB has recently demonstrated superior performance \citep{eimer23} and can be utilised for all three optimisation problems introduced in Section \ref{back:combined}. 
DEHB is a black-box, multi-fidelity hyperparameter optimisation method combining differential evolution \citep{storn96} and HyperBand \citep{li18}. 
Its multi-fidelity approach involves running numerous parameter configurations with a limited budget (e.g., a fraction of training steps) and advancing promising configurations to the next higher budget. 
This strategy allows for efficient exploration of the parameter space by testing a large number of configurations while avoiding wasteful evaluations on suboptimal configurations.
The best-performing parameter configuration observed during optimisation is called the \emph{incumbent} configuration.

\section{Related Work}

Our work aims to optimise RL algorithms by focusing jointly on hyperparameters and reward shapes to 
consistently obtain policies with high performance.
The critical importance of hyperparameter tuning in deep RL is well-established \citep{andrychowicz20, henderson18, islam17}. 
Similarly, reward shaping is recognised as important for fast and stable training \citep{ng99, gupta22}. 
The development of stable and reliable policies has been explored in risk-averse, multi-objective RL \citep{garcia15, la13}, employing a straightforward variance-based performance penalty among many possible methods.

For black-box hyperparameter and reward shape optimisation, several methods have already been developed in the framework of AutoRL \citep{parker20}, a data-driven approach for systematically optimising RL algorithms through automated machine learning.
However, these methods only target either the hyperparameter or reward-shape optimisation problem.
Black-box methods optimising hyperparameters comprise  population-based \citep{jaderberg19, parker20, wan22} and multi-fidelity methods \citep{falkner18, awad21}.
A recent study \citep{eimer23} highlights the effectiveness of the multi-fidelity DEHB procedure for RL.
Black-box methods for optimising reward shapes have been studied using evolutionary methods \citep{zheng18, faust19, wang22}.
None of the mentioned works for AutoRL consider joint optimisation of hyperparameters and reward parameters.

The differences in performance achieved by separate hyperparameter optimisation and reward weight optimisations have been discussed by \citet{faust19}, showing that reward parameters alone can improve performance and search efficiency compared to hyperparameter tuning.
To the best of our knowledge, no comprehensive investigation has been conducted into whether combined reward and hyperparameter optimisation is generally possible and examined its potential in depth.
Moving beyond this, \citet{jaderberg19} provided initial evidence that joint optimisation of hyperparameters and reward shape can outperform standard hyperparameter optimisation with manual reward shaping. 
However, their findings were limited to a single environment and presented as a custom solution, focusing solely on solving one specific environment.
Furthermore, they did not thoroughly investigate the effects of a combined optimisation approach.

\section{Setup of Experiments}

In this section, we describe the setup of our experiments, the results of which will be discussed in Section \ref{sec:results}.
The experiments detailed in Section \ref{sec:inted_exp} aim to examine the relationship between specific hyperparameters and reward weights to better understand their interdependencies and the necessity of joint optimisation. 
Subsequently, the experiments in Section \ref{sec:optimisation} empirically investigate the performance of joint optimisation compared to individual optimisation to analyse differences in performance and cost.
We trained PPO and SAC agents in four environments, each with a specific task objective: 
in Gymnasium's continuous LunarLander \citep{gym23}, a probe aims to minimise landing time; in Google Brax Ant and Humanoid \citep{brax21}, a walking robot aims to maximise travel distance; and in Robosuite Wipe \citep{robosuite20}, a simulated robot arm seeks to maximise the amount of dirt wiped from a table. 
All environments were chosen for non-trivial reward structures and for posing difficult hyperparameter optimisation problems.
Specifically, Humanoid is notoriously difficult to solve, and the Wipe environment has a large reward parameter space that needs to be optimised.
The Wipe environment represents a task that has not yet been extensively studied in the literature and is closely related to real-world applications. 
This allows us to test the applicability of our combined optimisation approach to environments that are less well-established in the field, yet of high practical interest.
More information about the environments and their reward structure can be found in Appendix~\ref{app:environments}.
For training with LunarLander and Wipe, we employed the stable-baseline's Jax PPO and SAC implementation \citep{raffin21}, while for the Google Brax environments, we utilised the Google Brax GPU implementations. 
Implementation details can be found in the supplementary code repository \url{https://github.com/ADA-research/combined_hpo_and_reward_shaping}.

\subsection{Interdependency of Hyperparameters and Reward Parameters}
\label{sec:inted_exp}

We conducted an exhaustive landscape analysis for PPO training on LunarLander, exploring pairwise combinations of a hyper- and reward parameter to better understand their interdependencies and substantiate the intuition that both components should be optimised jointly.
The parameters not considered in each pair were fixed to their baseline value.
A resolution of 100 values per parameter was applied, and the training performance of each pair was measured by the single-objective task performance and averaged over 10 seeds.
In terms of hyperparameters, we considered the discount factor $\gamma$, generalised advantage estimation $\lambda$, and learning rate $\eta$,
and in terms of reward parameters, the tilting, distance, and velocity weight.
A logarithmic grid was applied to the discount factor and learning rate, with points positioned at equidistant logarithms. 
A uniform grid of equidistant points was applied to all other hyper- and reward parameters.
Both choices were also used in our later optimisation experiments.

\subsection{Optimisation of Hyperparameters and Reward Parameters}
\label{sec:optimisation}

We conducted optimisation experiments to empirically compare the performance of joint optimisation with individual optimisation of hyperparameters and reward parameters;
our goal was to understand the practicality of joint optimisation in finding well-performing hyperparameters and reward parameters without requiring any manual tuning.

We used the black-box algorithm DEHB for the three optimisation problems introduced in Section \ref{back:combined}.
The hyperparameter search spaces for PPO and SAC consist of four and seven parameters, respectively, that are commonly optimised and known to impact performance significantly.
In particular, learning rate and discount factor were optimised for PPO and SAC.
The hyperparameters not included in the search space were fixed at the baseline values of each training.
For the reward function, we adjusted the weight parameters of each environment's reward shape. 
LunarLandar has four reward parameters, Ant and Humanoid three, and Wipe seven.
The hyperparameters not optimised in reward-weight-only optimisation were set to the algorithm's training baseline values for the respective environment.
The reward parameters not optimised in the case of hyperparameter-only optimisation were set to the default values of the respective environments.
In the combined optimisation approaches, all hyperparameters and reward parameters in the search space were optimised from scratch.
The search spaces and baseline values for hyperparameters are detailed in Appendix \ref{app:search}, while the search spaces and default values for reward parameters are provided in Appendix \ref{app:environments}.
DEHB has been demonstrated to outperform random search for hyperparameter optimisation \citep{awad21}. 
To analyse its effect on the optimisation of reward parameters, we used a random search approach for the  combined optimisation task, where the hyperparameters are optimised with DEHB, but the reward parameters are chosen randomly in each optimisation step. 

In our setup, we set the fidelity of DEHB to equal the number of RL training steps.
DEHB evaluates parameter configurations during the optimisation using three training step budgets, each increasing by a factor of three, with the largest matching the baseline's training steps.
The fitness of each configuration is determined by the average performance metric after training with the designated steps over three random seeds.
We performed experiments using the single- and multi-objective task objective performance metrics introduced in Section \ref{back:combined}.
Each optimisation experiment was conducted with five random seeds, and 
each resulting final incumbent configuration was evaluated by training using ten additional random seeds and evaluating performance on the corresponding task objective.

SAC is particularly sensitive to the scaling of the reward signal, since it influences the agent's exploration behaviour \citep{haarnoja18}. 
The reward scale $\alpha$ has only been optimised as part of the SAC baseline for Humanoid and Ant.
Thus, we separately optimised the reward scale $\alpha$ only for Humanoid and Ant and kept the reward scale fixed to $\alpha = 1$ for the other environments.
Details on how reward scaling was performed can be found in Appendix \ref{app:scaling}.

The overall optimisation budget for DEHB with PPO and SAC equals 133 and 80 full training step budgets, respectively.
Due to its computational demands, for the Wipe environment, we only considered SAC. 
The wall-clock times for the PPO and SAC optimisations are about \SI{4}{\hour} and \SI{60}{\hour} for LunarLander, \SI{12}{\hour} and \SI{15}{\hour} for Ant, \SI{36}{\hour} and \SI{60}{\hour} for Humanoid, and \SI{120}{\hour} for Wipe.
An overview of our execution environment and the overall computational cost can be found in Appendix \ref{app:exc_environment}.

\section{Empirical Results}
\label{sec:results}

We now present the results from our experiments.
First, we show the complex interdependencies between hyperparameters and reward weights. 
Second, we demonstrate that joint optimisation can match or outperform individual optimisation and produce policies with substantially lower variance.

\subsection{Interdependency between Hyperparameters and Reward Parameters}

\begin{wrapfigure}[30]{R}{0.6\textwidth}
    \includegraphics[width=0.6\textwidth]{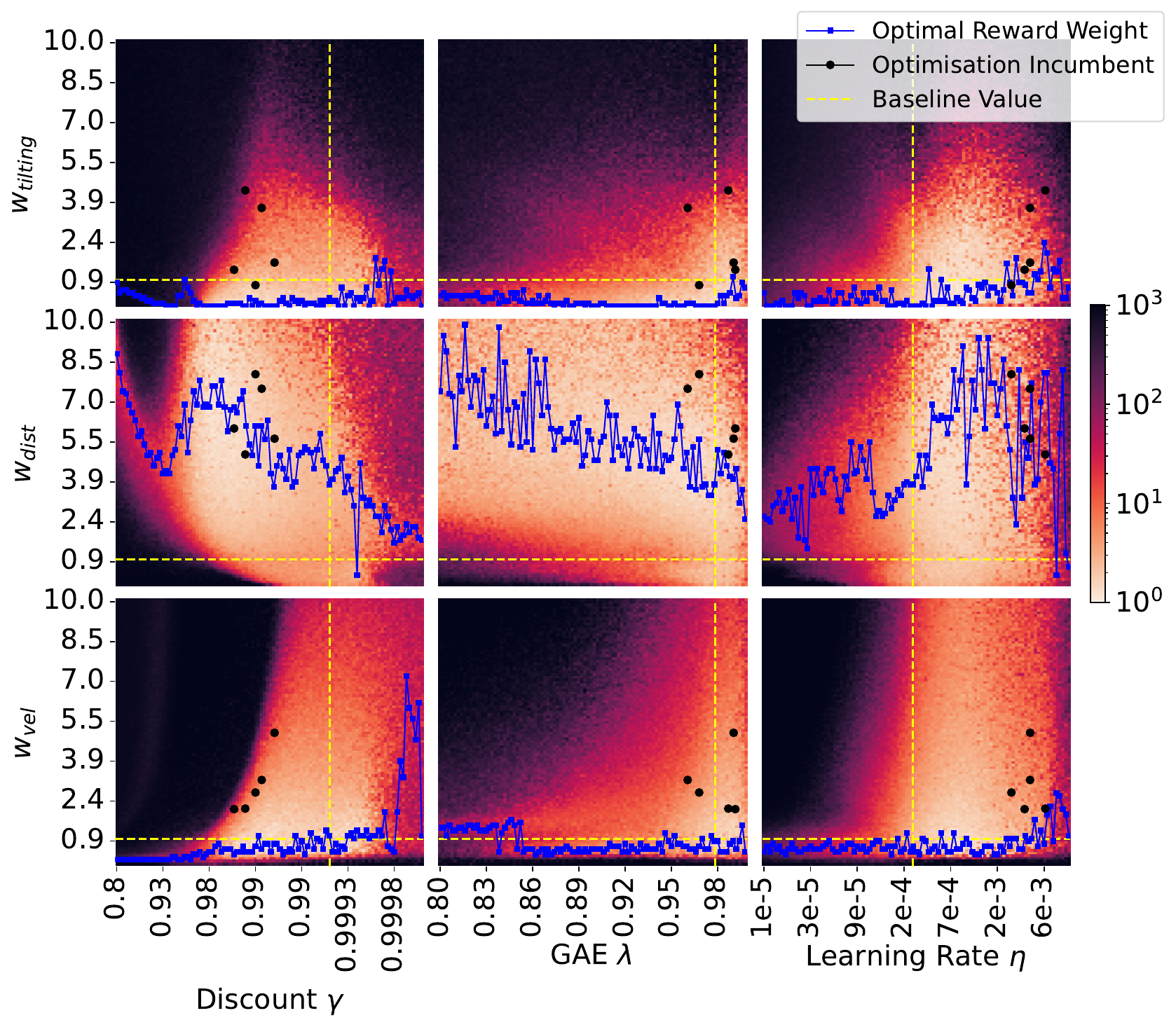}
    \caption{
        Landscapes depicting the average return on LunarLander for pairwise hyperparameters and reward weights over ten PPO trainings.
        Lower values (lighter) correspond to faster landing (better performance). 
        Yellow lines mark each parameter's default value.
        Blue lines denote the best-performing reward weights for each hyperparameter value.
        The black dots mark the incumbent configurations found in the joint optimisation experiments in Section \ref{sec:joint_opt_perf}.
    }
    \label{fig:heatmap_rewmin}
\end{wrapfigure}

The parameter landscapes for LunarLander with PPO are shown in Figure \ref{fig:heatmap_rewmin}. 
We observe an interdependency of varying strength between the hyperparameters and reward parameters.
In all cases, the behaviour of specific reward parameters changes with different hyperparameter values.
The ranges for reward parameters that lead to successful training vary depending on the hyperparameter and exhibit sharp boundaries in some cases.
In particular, we observed ranges of reward parameters where performance deteriorates across all possible hyperparameter values.

Regarding the relation between hyperparameters and best-performing reward parameters (indicated by the blue lines in the plot), we observed a strong dependency for the distance weight and weaker dependencies for the velocity weights. 
In particular, we see a non-convex region of successful training parameters for the distance weight.
Furthermore, we see large changes of the distance weight in its optimisation space.
The tilting weight shows almost no dependency on the hyperparameters.
Finally, our landscapes suggest that optimal value of the tilting weight is mostly near zero, which suggests that it is mostly irrelevant to RL training.
In addition, we observed that the incumbent configurations in the joint optimisation experiments for LunarLander, presented in Section \ref{sec:joint_opt_perf} (indicated by the black dots in the plot), are often not fully located in high-performing regions. 
We believe this is due to the larger search space during optimisation, which introduces additional dependencies on other parameters that impact performance in these regions. 
This further highlights the interdependencies of the parameters within the context of the full search space.
In Appendix \ref{app:interdependency}, we report landscapes showing the optimal hyperparameters with respect to the reward parameters, showing similar dependencies.

Overall, our results 
indicate that hyperparameters and reward parameters are interdependent and that finding high-performing hyperparameters necessitates well-chosen reward parameters and vice versa. 
This confirms the intuition this work is based on: optimising the hyperparameters and reward shape should not be considered independently but instead approached jointly.

\subsection{Joint Optimisation Performance}
\label{sec:joint_opt_perf}

Table \ref{tab:external_objective} reports the results of our optimisation experiments.
Performance is shown in terms of single-objective task performance and the coefficient of variation (in percent).
As outlined in Section \ref{sec:optimisation}, each experiment consists of five optimisation runs, with the incumbent parameter configuration of each run evaluated through ten RL training runs. 
The performance results in Table \ref{tab:external_objective} are derived by calculating the median performance for each optimisation run across its ten evaluations and then computing the median of these five values for each experiment.
The median coefficients of variation are calculated analogously.
We chose the median over the mean to present our results, as it is more robust to outliers.
To gain further insights into the statistical differences between the optimisation experiments, we employed linear mixed-effects model analysis \citep{gelman2006}. 
For each combination of environment and algorithm, we performed pairwise comparisons of the aggregated 50 evaluation runs of the best-performing experiment with those of the related optimisation experiments.
The mixed-effects model allows us to test for statistically significant differences in the results of two optimisation experiments, using all available data, while correctly handling the dependencies between optimisation runs. 
We show the best performance and all results that show no statistically significant differences (at significance level 0.05) to it in boldface.
Details on how the test was conducted and its assumptions can be found in Appendix \ref{app:lme_test}.
Boxplots of the median performance from the five optimisation runs for each experiment as well as boxplots of the 50 aggregated evaluations across all optimisation and training runs are presented in Appendix \ref{app:boxplots}.

\begin{table}[tb]
    \centering
    \small
\renewcommand{\arraystretch}{1.1}
\begin{tabular}{|c|cc|cccc|}
\hline
\multirow{3}{*}{Environment}                                                                  & \multicolumn{1}{c|}{\multirow{3}{*}{HPO}} & \multirow{3}{*}{RPO} & \multicolumn{4}{c|}{Task Performance $\mathbb{E}_{\substack{\tau \sim \pi}}[o_\text{goal}(\tau)]\ (100 \cdot \text{CV}_{\tau \sim \pi}[o_\text{goal}(\tau)]$)}                                                                                                                                 \\ \cline{4-7} 
                                                                                              & \multicolumn{1}{c|}{}                     &                      & \multicolumn{2}{c|}{PPO}                                                                     & \multicolumn{2}{c|}{SAC}                                                 \\ \cline{4-7} 
                                                                                              & \multicolumn{1}{c|}{}                     &                      & \multicolumn{1}{c|}{Single Obj.}              & \multicolumn{1}{c|}{Multi Obj.}              & \multicolumn{1}{c|}{Single Obj.}              & Multi Obj.               \\ \hline\hline
\multirow{5}{*}{\begin{tabular}[c]{@{}c@{}}Gymnasium\\ LunarLander\\ (minimise)\end{tabular}} & \multicolumn{2}{c|}{base}                                        & \multicolumn{2}{c|}{273 (11\%)}                                                          & \multicolumn{2}{c|}{208 (27\%)}                                      \\ \cline{2-7} 
                                                                                              & \multicolumn{1}{c|}{base}                 & DEHB                 & \multicolumn{1}{c|}{287 (31\%)}           & \multicolumn{1}{c|}{\textbf{223 (10\%)}} & \multicolumn{1}{c|}{\textbf{175 (14\%)}}  & \textbf{174 (13\%)}  \\ \cline{2-7} 
                                                                                              & \multicolumn{1}{c|}{DEHB}                 & base                 & \multicolumn{1}{c|}{265 (27\%)}           & \multicolumn{1}{c|}{277 (11\%)}          & \multicolumn{1}{c|}{194 (23\%)}           & 186 (15\%)           \\ \cline{2-7} 
                                                                                              & \multicolumn{1}{c|}{DEHB}                 & RS                   & \multicolumn{1}{c|}{262 (38\%)}           & \multicolumn{1}{c|}{252 (24\%)}          & \multicolumn{1}{c|}{\textbf{171 (15\%)}}  & 193 (18\%)           \\ \cline{2-7} 
                                                                                              & \multicolumn{2}{c|}{DEHB (ours)}                                 & \multicolumn{1}{c|}{\textbf{234 (25\%)}}  & \multicolumn{1}{c|}{\textbf{227 (15\%)}} & \multicolumn{1}{c|}{\textbf{177 (23\%)}}  & \textbf{182 (21\%)}  \\ \hline\hline
\multirow{5}{*}{\begin{tabular}[c]{@{}c@{}}Google Brax \\ Ant\\ (maximise)\end{tabular}}      & \multicolumn{2}{c|}{base}                                        & \multicolumn{2}{c|}{6785 (16\%)}                                                         & \multicolumn{2}{c|}{8054 (28\%)}                                     \\ \cline{2-7} 
                                                                                              & \multicolumn{1}{c|}{base}                 & DEHB                 & \multicolumn{1}{c|}{6706 (17\%)}          & \multicolumn{1}{c|}{6663 (14\%)}         & \multicolumn{1}{c|}{7927 (32\%)}          & 7994 (29\%)          \\ \cline{2-7} 
                                                                                              & \multicolumn{1}{c|}{DEHB}                 & base                 & \multicolumn{1}{c|}{\textbf{8111 (14\%)}} & \multicolumn{1}{c|}{7842 (6\%)}          & \multicolumn{1}{c|}{\textbf{8282 (21\%)}} & \textbf{8216 (13\%)} \\ \cline{2-7} 
                                                                                              & \multicolumn{1}{c|}{DEHB}                 & RS                   & \multicolumn{1}{c|}{8013 (16\%)}          & \multicolumn{1}{c|}{-}                       & \multicolumn{1}{c|}{\textbf{8064 (21\%)}} & -                        \\ \cline{2-7} 
                                                                                              & \multicolumn{2}{c|}{DEHB (ours)}                                        & \multicolumn{1}{c|}{8049 (12\%)}          & \multicolumn{1}{c|}{\textbf{7923 (6\%)}} & \multicolumn{1}{c|}{\textbf{8199 (23\%)}} & \textbf{8169 (18\%)} \\ \hline\hline
\multirow{5}{*}{\begin{tabular}[c]{@{}c@{}}Google Brax \\ Humanoid\\ (maximise)\end{tabular}} & \multicolumn{2}{c|}{base}                                        & \multicolumn{2}{c|}{4196 (<1\%)}                                                          & \multicolumn{2}{c|}{3273 (11\%)}                                     \\ \cline{2-7} 
                                                                                              & \multicolumn{1}{c|}{base}                 & DEHB                 & \multicolumn{1}{c|}{4464 (<1\%)}           & \multicolumn{1}{c|}{4472 (<1\%)}          & \multicolumn{1}{c|}{5284 (11\%)}          & 5208 (8\%)           \\ \cline{2-7} 
                                                                                              & \multicolumn{1}{c|}{DEHB}                 & base                 & \multicolumn{1}{c|}{4826 (1\%)}           & \multicolumn{1}{c|}{4719 (<1\%)}          & \multicolumn{1}{c|}{4881 (18\%)}          & 4466 (15\%)          \\ \cline{2-7} 
                                                                                              & \multicolumn{1}{c|}{DEHB}                 & RS                   & \multicolumn{1}{c|}{5112 (2\%)}           & \multicolumn{1}{c|}{-}                       & \multicolumn{1}{c|}{\textbf{5913 (17\%)}} & -                        \\ \cline{2-7} 
                                                                                              & \multicolumn{2}{c|}{DEHB (ours)}                                 & \multicolumn{1}{c|}{\textbf{5433 (7\%)}}  & \multicolumn{1}{c|}{\textbf{5485 (1\%)}} & \multicolumn{1}{c|}{\textbf{6033 (12\%)}} & \textbf{6103 (1\%)}  \\ \hline\hline
\multirow{5}{*}{\begin{tabular}[c]{@{}c@{}}Robosuite \\ Wipe\\ (maximise)\end{tabular}}       & \multicolumn{2}{c|}{base}                                        & \multicolumn{2}{c|}{\multirow{5}{*}{-}}                                                      & \multicolumn{2}{c|}{101 (24\%)}            \\ \cline{2-3} \cline{6-7}
                                                                                              & \multicolumn{1}{c|}{base}                 & DEHB                 & \multicolumn{2}{c|}{}                                                                        & \multicolumn{1}{c|}{108 (24\%)}           &  \multicolumn{1}{c|}{114 (20\%)}  \\ \cline{2-3} \cline{6-7}
                                                                                              & \multicolumn{1}{c|}{DEHB}                 & base                 & \multicolumn{2}{c|}{}                                                                        & \multicolumn{1}{c|}{132 (10\%)}          &  \multicolumn{1}{c|}{\textbf{131 (11\%)}}    \\ \cline{2-3} \cline{6-7}
                                                                                              & \multicolumn{1}{c|}{DEHB}                 & RS                   & \multicolumn{2}{c|}{}                                                                        & \multicolumn{1}{c|}{134 (10\%)}          &  \multicolumn{1}{c|}{-}    \\ \cline{2-3} \cline{6-7}
                                                                                              & \multicolumn{2}{c|}{DEHB (ours)}                                 & \multicolumn{2}{c|}{}                                                                        & \multicolumn{1}{c|}{\textbf{136 (8\%)}}   &  \multicolumn{1}{c|}{\textbf{130 (10\%)}}   \\ \hline
\end{tabular}

    \caption{
        Median performance for our optimisation experiments. 
        \textit{HPO} and \textit{RPO} show the optimisation method for hyper- and reward parameters: \textit{base} for fixing to baseline values, \textit{DEHB} and \textit{RS} for optimisation with DEHB or random search. 
        Each environment's first row is baseline performance, followed by optimising reward-, hyperparameters, or both. 
        Best performance are highlighted in bold for each environment and column (multiple bold entries mark statistically insignificant differences).
    }
    \label{tab:external_objective}
\end{table}

Our results show that simultaneously optimising hyperparameters and reward parameters consistently matches or outperforms individual optimisation, without depending on tuned baseline parameters for non-optimised components. 
The only outlier is the single-objective PPO Ant optimisation.
Significant performance gains are observed in the complex Humanoid and Wipe environments, while the simpler Ant and LunarLander environments, which are mostly solved using baseline parameter settings, generally show no additional improvements from joint optimisation.
However, even if performance only matches the baseline parameters, joint optimisation still offers the advantage of not requiring hand-tuning, while addressing the mutual dependencies of hyperparameter and reward parameters.

In LunarLander, Ant, and Humanoid, the optimised incumbent of our joint optimisation on the respective default reward function generally achieves performance considered to solve the environment.
For Robosuite Wipe, the average objective score comes close to the maximum of 142.
Especially in our experiments with LunarLander, Humanoid and Wipe, the policy improvements could be seen not only in the improved average objective score, but also in qualitative improvements in the agents' behaviour
(representative videos can be found in the supplementary material).
Therefore, combined optimisation shows competitive performance for already well-studied environments as well as the less-studied Wipe environment.
We do not observe a clear pattern in Table \ref{tab:external_objective} that indicates whether optimising solely hyperparameters or reward parameters consistently outperforms the other. 
This underscores the necessity of joint optimisation to automatically determine which component requires more optimisation, especially for novel environments, where prior knowledge about the dynamics is lacking.
Unsurprisingly, DEHB outperforms random search in almost all our experiments.
We report the best hyperparameter and reward parameter values for each environment and algorithm in Appendix \ref{app:best_configurations}.
Appendix \ref{app:default_shape_results} provides the results of policies obtained for each configuration when evaluated using the default shaped reward function of the respective environment. 

\begin{figure}[tb]
    \centering
    \includegraphics[width=1.0\textwidth]{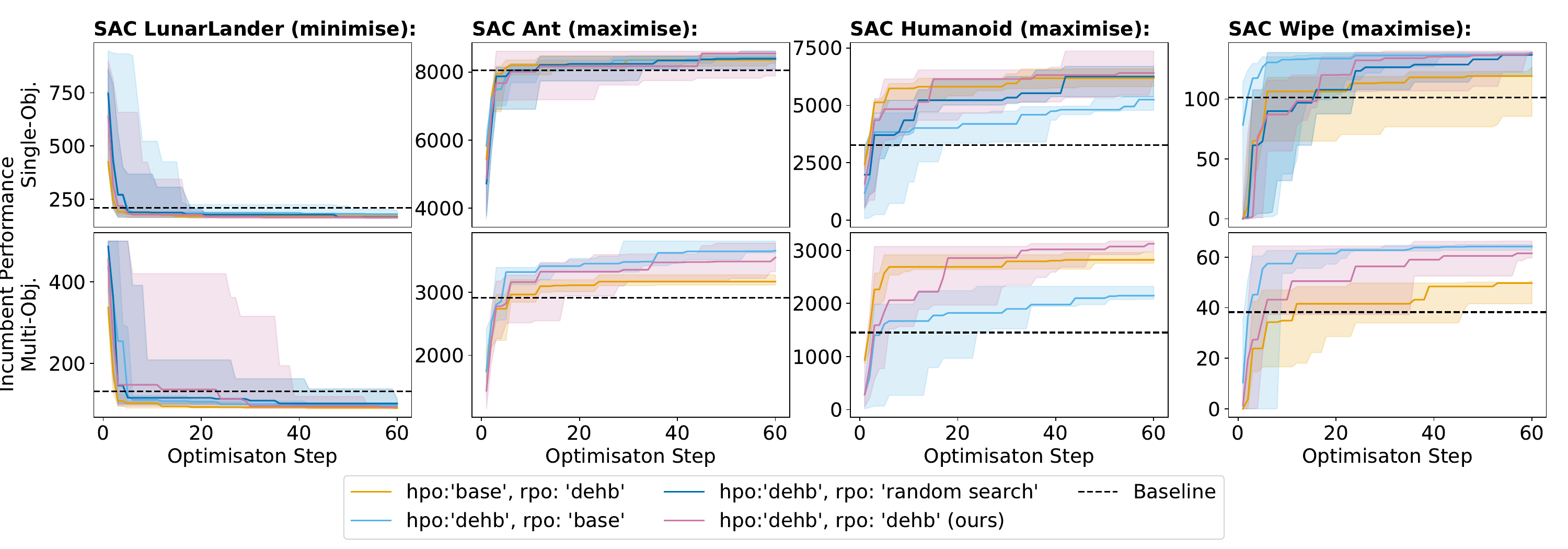}
    \caption{
    Incumbent performance in terms of median optimisation objective 
    across the five optimisation runs for the SAC experiments at each time step;
    shaded areas indicate min and max values. 
    The performance drop in the multi-objective experiments is due to the weighted penalty term.
    }
    \label{fig:incumb}
\end{figure}

In Figure \ref{fig:incumb}, we show median incumbent performance during our SAC experiments at each optimisation time step. 
The speed of the combined optimisation is comparable to that of the individual optimisation approaches, despite involving much larger search spaces. 
In all environments except multi-objective Wipe, combined optimisation already matches the performance of the best-performing individual optimisation after roughly a third of the total optimisation steps and continues to improve. 
For multi-objective Wipe, this is achieved after two-thirds of the total optimisation steps.
Similar trends are observed for the PPO results, shown in Appendix \ref{app:best_configurations}. 
This indicates that combined optimisation, despite the larger search space, requires minimal additional computational effort in terms of optimisation time.

\subsection{Single- \emph{vs} Multi-objective Optimisation}

From Table \ref{tab:external_objective}, we conclude that multi-objective optimisation can improve policy stability by including a penalty for large standard deviation in performance.
These improvements come with only marginal performance loss and sometimes even achieve slight gains;
this is the case for Ant, Humanoid and LunarLander, where, in particular, for Humanoid and PPO LunarLander, improved performance is achieved.
Only for Wipe, we observed that stability is not further improved compared to the single-objective training.
RL is notoriously sensitive to hyperparameter settings.
Therefore, optimising hyperparameters using a variance penalty for newly developed algorithms or novel scenarios can lead to increased stability and thus greatly facilitate research and applications.

\section{Conclusions and Future Work}

In this work, we have demonstrated the importance of 
jointly optimising hyperparameters and reward parameters.
We illustrated dependencies in a simple environment, highlighting the circular dependency encountered in optimising hyperparameters and reward parameters and underscoring the need for simultaneous optimisation.
Our empirical results indicate that this joint optimisation is feasible and can match or surpass the performance of individual optimisation approaches without requiring separate parameter tuning for the non-optimised component.
Additionally, we demonstrated that this approach requires minimal extra computational effort and is applicable to practical environments, not yet extensively studied.
We conclude that combined optimisation should be the best practice for RL optimisation.
While we have focused on optimising specific reward parameters within a predefined reward structure, future work should explore a broader range of reward function combinations.
Such an extension could consider further aspects of the reward function, including metrics, exponentiation, or specific functional choices, such as nonlinear transformations.

Our results further indicate that including a variance penalty in a multi-objective optimisation can substantially enhance the  performance variance of a given policy, with little or no reductions in performance.
This emphasises the value of combined optimisation in achieving a good balance between a high average objective score and achieving this performance consistently. 
This improvement in stability is often a crucial requirement in reinforcement learning, enhancing reproducibility and the reliability of results in varying environments.
Future research should investigate more sophisticated risk-averse metrics and thoroughly assess the trade-off between a policy's performance and stability.

\subsubsection*{Acknowledgments}
\label{sec:ack}

The authors would like to thank Theresa Eimer for helpful discussions regarding the search space design, Anna Münz for help with statistical tests, and Anja Jankovic for helpful feedback.
We gratefully acknowledge computing resources provided by the NHR Center NHR4CES at RWTH Aachen University (p0021208), funded by the Federal Ministry of Education and Research, and the state governments participating on the basis of the resolutions of the GWK for national high performance computing at universities.
This research was supported in part by an Alexander von Humboldt Professorship in AI held by HH and by the ``Demonstrations- und Transfernetzwerk KI in der Produktion (ProKI-Netz)'' initiative, funded by the German Federal Ministry of Education and Research (BMBF, grant number 02P22A010).

\bibliography{main}
\bibliographystyle{rlc}

\newpage

\appendix

\section{Environments and Reward Parameter Search Spaces}
\label{app:environments}

In this section, we give a detailed overview of the environments used in our experiments and the reward parameters we are optimising in each environment's reward function.
The default values and respective search spaces of the reward parameters can be found in Table \ref{tab:rweight_params_search}.
We opted not to optimise the terminal rewards $r$ for the LunarLander and Wipe environments, as these constitute the sparse rewards that are addressed through the optimisation of the reward shape.

For mapping a reward weight $w_{i}$ to its search space, we always used the mapping $w_{i} \mapsto [0, 10^{n}]$, where $n \in \mathbb{N}_{0}$ is the smallest integer such that $w_{i} < 10^{n}$;
we chose this approach, since it preserves the general magnitude of the reward parameters, while relying less on their initial ratios.
We believe that practitioners typically have a rough idea about the importance of different components but find it difficult to obtain the exact ratios between them. 

\begin{table}[tb]
    \centering
    \footnotesize
\begin{tabular}{|l|cc|cc|cc|cc|}
\hline
\multirow{2}{*}{Reward Weight} & \multicolumn{2}{c|}{LunarLander}                                                                                                      & \multicolumn{2}{c|}{Ant}                                                                                                              & \multicolumn{2}{c|}{Humanoid}                                                                                                         & \multicolumn{2}{c|}{Wipe}                                                                                                             \\ \cline{2-9} 
                               & \multicolumn{1}{c|}{\begin{tabular}[c]{@{}c@{}}Default\\ Value\end{tabular}} & \begin{tabular}[c]{@{}c@{}}Search\\ Space\end{tabular} & \multicolumn{1}{c|}{\begin{tabular}[c]{@{}c@{}}Default\\ Value\end{tabular}} & \begin{tabular}[c]{@{}c@{}}Search\\ Range\end{tabular} & \multicolumn{1}{c|}{\begin{tabular}[c]{@{}c@{}}Default\\ Value\end{tabular}} & \begin{tabular}[c]{@{}c@{}}Search\\ Space\end{tabular} & \multicolumn{1}{c|}{\begin{tabular}[c]{@{}c@{}}Default\\ Value\end{tabular}} & \begin{tabular}[c]{@{}c@{}}Search\\ Space\end{tabular} \\ \hline
$w_\text{dist}$                & \multicolumn{1}{c|}{$100$}                                                   & $[0, 1000]$                                            & \multicolumn{1}{c|}{$1$}                                                     & $[0, 10]$                                              & \multicolumn{1}{c|}{$1.25$}                                                  & $[0, 10]$                                              & \multicolumn{1}{c|}{$5$}                                                     & $[0, 10]$                                              \\ \hline
$w_\text{dist\_th}$            & \multicolumn{2}{c|}{-}                                                                                                                & \multicolumn{2}{c|}{\multirow{4}{*}{-}}                                                                                               & \multicolumn{2}{c|}{\multirow{4}{*}{-}}                                                                                               & \multicolumn{1}{c|}{$5$}                                                     & $[0, 10]$                                              \\ \cline{1-3} \cline{8-9} 
$w_\text{vel}$                 & \multicolumn{1}{c|}{$100$}                                                   & $[0, 1000]$                                            & \multicolumn{2}{c|}{}                                                                                                                 & \multicolumn{2}{c|}{}                                                                                                                 & \multicolumn{1}{c|}{$0$}                                                     & $[0, 1]$                                               \\ \cline{1-3} \cline{8-9} 
$w_\text{tilting}$             & \multicolumn{1}{c|}{$100$}                                                   & $[0, 1000]$                                            & \multicolumn{2}{c|}{}                                                                                                                 & \multicolumn{2}{c|}{}                                                                                                                 & \multicolumn{2}{c|}{-}                                                                                                                \\ \cline{1-3} \cline{8-9} 
$w_\text{contact}$             & \multicolumn{1}{c|}{$10$}                                                    & $[0, 100]$                                             & \multicolumn{2}{c|}{}                                                                                                                 & \multicolumn{2}{c|}{}                                                                                                                 & \multicolumn{1}{c|}{$0.01$}                                                  & $[0, 1]$                                               \\ \hline
$w_\text{healthy}$             & \multicolumn{2}{c|}{\multirow{4}{*}{-}}                                                                                               & \multicolumn{1}{c|}{$1$}                                                     & $[0, 10]$                                              & \multicolumn{1}{c|}{$5$}                                                     & $[0, 10]$                                              & \multicolumn{2}{c|}{-}                                                                                                                \\ \cline{1-1} \cline{4-9} 
$w_\text{unhealthy}$           & \multicolumn{2}{c|}{}                                                                                                                 & \multicolumn{2}{c|}{-}                                                                                                                & \multicolumn{2}{c|}{-}                                                                                                                & \multicolumn{1}{l|}{$-10$}                                                   & \multicolumn{1}{l|}{$[-100, 0]$}                       \\ \cline{1-1} \cline{4-9} 
$w_\text{force}$               & \multicolumn{2}{c|}{}                                                                                                                 & \multicolumn{1}{c|}{$0.5$}                                                   & $[0, 1]$                                               & \multicolumn{1}{c|}{$0.1$}                                                   & $[0, 1]$                                               & \multicolumn{1}{c|}{$0.05$}                                                  & $[0, 1]$                                               \\ \cline{1-1} \cline{4-9} 
$w_\text{wiped}$               & \multicolumn{2}{c|}{}                                                                                                                 & \multicolumn{2}{c|}{-}                                                                                                                & \multicolumn{2}{c|}{-}                                                                                                                & \multicolumn{1}{c|}{$50$}                                                    & $[0, 100]$                                             \\ \hline
\end{tabular}
    \caption{
        Default reward weights for each environment and the corresponding search spaces of our optimisation experiments.
    }
    \label{tab:rweight_params_search}
\end{table}

\begin{figure}[tb]
    \centering
    \includegraphics[width=\textwidth]{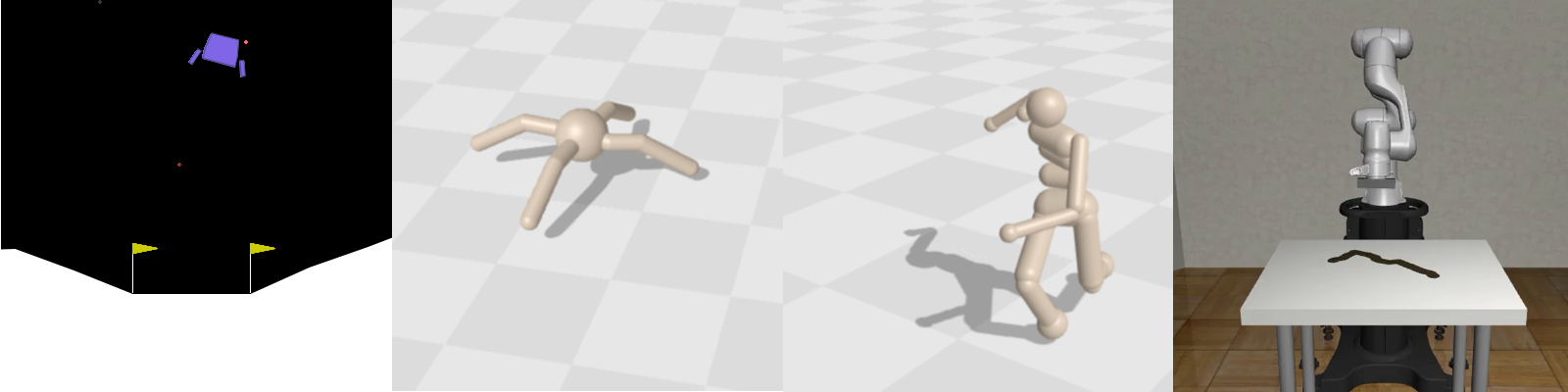}
    \caption{Illustrations from left to right of the environments Gymnasium LunarLander, Google Brax Ant and Humanoid, and Robosuite Wipe.}
    \label{fig:envs}
\end{figure}

\subsection{Gymnasium LunarLander:}
The objective of the environment is to navigate a probe to a designated landing platform safely.
We considered the environment's variant with continuous control inputs.
In the shaped reward, positive rewards are given for moving closer to the landing platform and negative rewards for moving further away.
A positive reward is granted for making successful contact with the platform using the probe's legs.
Negative rewards are imposed for high velocities and tilting the probe excessively.
Fuel consumption by the probe's engine, when activated, results in negative rewards. 
However, we considered fuel consumption as a constant physical attribute of the probe and did not consider it in our optimisations.
The overall shaped reward function is given by
\begin{align}
    \tilde{r}^{\alpha, w} := \alpha\cdot(w_{\text{dist}}\cdot\tilde{r}_{\text{dist}} + w_\text{contact}\cdot\tilde{r}_\text{contact} - w_\text{vel}\cdot\tilde{r}_\text{vel}  -  w_\text{tilting}\cdot\tilde{r}_\text{tilting} - \tilde{r}_\text{fuel} + r_{terminal}),
\end{align}
with $r_{terminal}$ being the environment's sparse reward signal of successful landing or crashing.

\subsection{Google Brax Ant and Humanoid:}
The task in both environments is to train a robot to walk forward in a specified direction.
In Ant, the robot is designed to resemble a four-legged ant and is human-like in Humanoid.
The environment's rewards consist of positive rewards for staying healthy (being able to continue walking) and a reward for the distance travelled in each timestep.
A negative reward for exercising large forces on the robot's joints is obtained.
The overall shaped reward function is given by
\begin{align}
    \tilde{r}^{\alpha, w} := \alpha\cdot(w_{\text{dist}}\cdot\tilde{r}_{\text{dist}} + w_{\text{healthy}}\cdot\tilde{r}_{\text{healthy}} - w_{\text{force}}\cdot\tilde{r}_{\text{force}}).
\end{align}

\subsection{Robosuite Wipe:}
The task is to wipe a table of dirt pegs with a simulated robot arm equipped with a sponge. 
Positive rewards are obtained for the sponge's distance to the dirt pegs, having contact with the table and exercising appropriate pressure on the table.
Negative rewards are obtained for applying excessive force on the table, large accelerations while moving or arm-limit collisions resulting in an unhealthy state.
The overall shaped reward function is given by
\begin{align}
    \tilde{r}^{\alpha, w} :=\ &\alpha\cdot(w_\text{wiped}\cdot\tilde{r}_\text{wiped} + w_\text{dist}\cdot(1 - \text{tanh}(w_\text{dist\_th}\cdot\tilde{r}_\text{dist})) + w_\text{contact}\cdot\tilde{r}_\text{contact} \\ &- w_\text{force}\cdot\tilde{r}_\text{force} - w_\text{vel}\cdot\tilde{r}_\text{vel} + w_\text{unhealthy}\cdot\tilde{r}_\text{unhealthy} + r_{terminal}),
\end{align}
with $r_{terminal}$ being the environment's sparse reward for wiping the table clean or not.

\section{Interdependecy between Hyperparameters and reward parameters}
\label{app:interdependency}

\begin{figure}[H]
    \centering
    \includegraphics[width=0.63\textwidth]{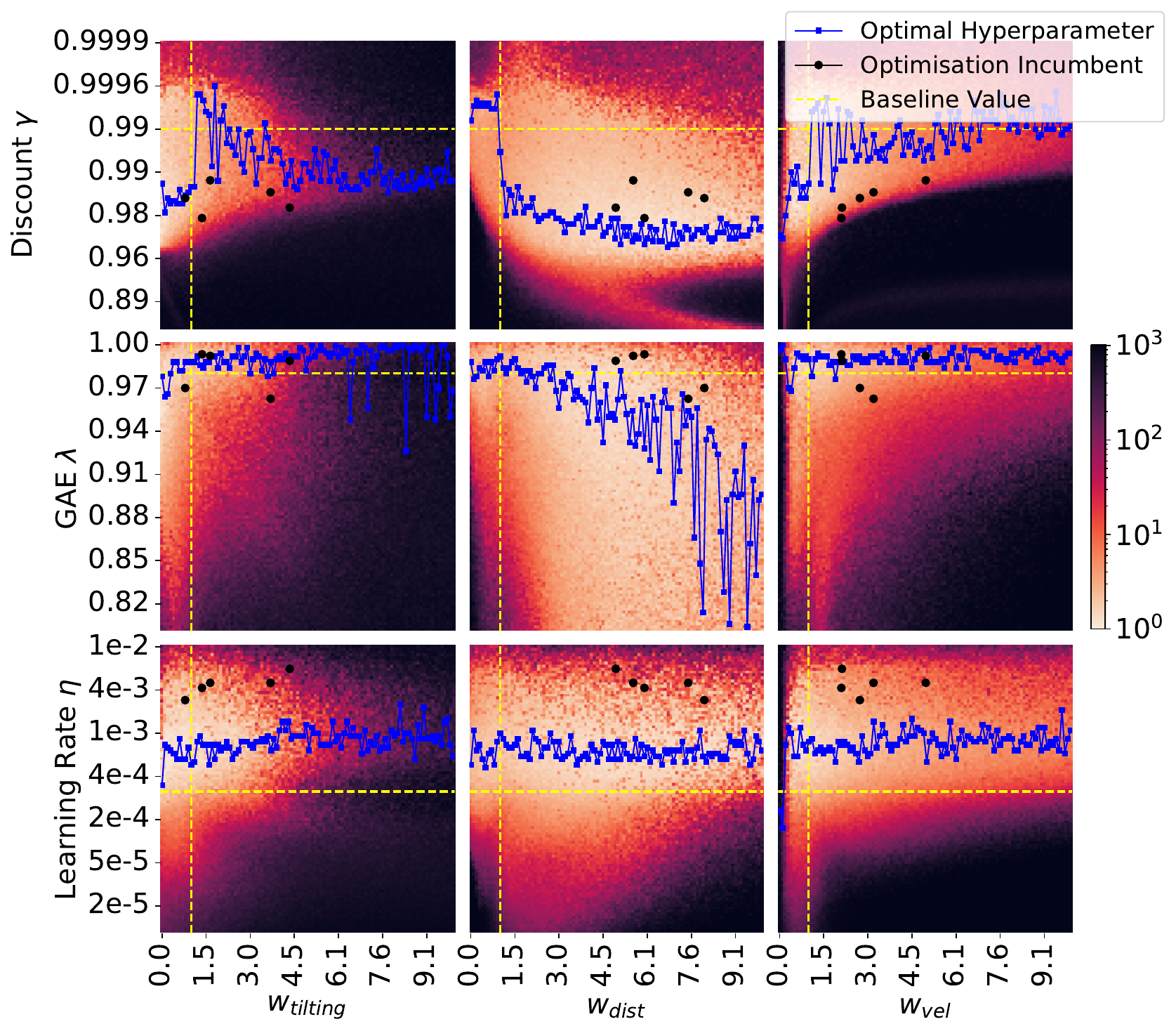}
    \caption{
        Landscapes depicting the average return on Gymnasium LunarLander for pairwise hyper- and reward parameters over ten PPO trainings.
        Lower values (lighter) correspond to faster landing time and, thus, better performance. 
        The yellow lines mark the default values for each parameter.
        The blue line denotes the best-performing hyperparameter value for each specific reward value.
        The black dots mark the incumbent configurations found in the joint optimisation experiments in Section \ref{sec:joint_opt_perf}
    }
    \label{fig:heatmap_algmin}
\end{figure}

In Figure \ref{fig:heatmap_algmin}, we present the same landscapes as in Figure \ref{fig:heatmap_rewmin} but mark the best-performing hyperparameter value for each reward weight (shown as the blue lines).
We can again observe strong dependencies and large changes in the best-performing hyperparameters for the discount factor and the generalised advantage estimate regarding the distance reward parameter. 
Only for the learning rate, we see almost no dependency on the reward parameters.

\section{Hyperparameter Baselines and Search Spaces}
\label{app:search}

In Table \ref{tab:ppo_baselines} and Table \ref{tab:sac_baselines}, we present the baselines and search spaces for PPO and SAC, respectively.
We reproduced all baselines in our setup and, in some cases, made slight modifications to improve their performance when possible.
The baselines for continuous LunarLander PPO and SAC have both been obtained from the stable-baselines 3 Zoo \citep{raffin20}.
The Google Brax Ant and Humanoid baselines are obtained from Brax's GitHub repository. 
For PPO, hyperparameters have been shared via a Google Colab notebook in the Google Brax GitHub repository.
For SAC, we utilise the performance results of a published hyperparameter sweep. 
In the case of Humanoid, we made small adjustments to the reported best-performing parameters in order to obtain the best results in our setup. 
Our adjustments align with the considered search space of their hyperparameter sweep.
We utilised the published baseline in \citet{robosuite20} for Robosuite Wipe with small adjustments to the training frequency and gradient steps to obtain better performance in our setup.

\begin{table}[tb]
    \centering
    \footnotesize
\begin{tabular}{|l|ccc|cc|}
\hline
\multirow{2}{*}{Hyperparameter} & \multicolumn{3}{c|}{Baseline Values}                                      & \multicolumn{2}{c|}{Search Space}                             \\ \cline{2-6} 
                                & \multicolumn{1}{c|}{LunarLander} & \multicolumn{1}{c|}{Ant}    & Humanoid & \multicolumn{1}{c|}{Range}             & Log-Scale            \\ \hline\hline
learning rate                   & \multicolumn{3}{c|}{$3\cdot10^{-4}$}                                      & \multicolumn{1}{c|}{$[1e^{-6}, 0.01]$} & \multirow{2}{*}{Yes} \\ \cline{1-5}
discounting                     & \multicolumn{1}{c|}{$0.999$}     & \multicolumn{2}{c|}{$0.97$}            & \multicolumn{1}{c|}{$[0.001, 0.02]$}   &                      \\ \hline
gae                             & \multicolumn{1}{c|}{$0.98$}      & \multicolumn{2}{c|}{$0.95$}            & \multicolumn{1}{c|}{$[0.8, 1.0]$}      & \multirow{5}{*}{No}  \\ \cline{1-5}
clipping rate                   & \multicolumn{1}{c|}{$0.2$}       & \multicolumn{2}{c|}{$0.3$}             & \multicolumn{1}{c|}{$[0.1, 0.4]$}      &                      \\ \cline{1-5}
entropy coef                    & \multicolumn{1}{c|}{$0.01$}      & \multicolumn{2}{c|}{$0.001$}           & \multicolumn{1}{c|}{$[0.0, 0.1]$}      &                      \\ \cline{1-5}
value coef                      & \multicolumn{3}{c|}{$0.5$}                                                & \multicolumn{1}{c|}{$[0.3, 0.7]$}      &                      \\ \cline{1-5}
batch size                      & \multicolumn{1}{c|}{$64$}        & \multicolumn{1}{c|}{$2048$} & $1024$   & \multicolumn{1}{c|}{$\{b/2, b, 2b\}$} &                      \\ \hline
training steps                  & \multicolumn{1}{c|}{$1e^6$}      & \multicolumn{2}{c|}{$5242880$}         & \multicolumn{2}{c|}{\multirow{6}{*}{-}}                       \\ \cline{1-4}
episode length                  & \multicolumn{3}{c|}{$1000$}                                               & \multicolumn{2}{c|}{}                                         \\ \cline{1-4}
num envs                        & \multicolumn{1}{c|}{$1$}         & \multicolumn{1}{c|}{$256$}  & $64$     & \multicolumn{2}{c|}{}                                         \\ \cline{1-4}
unroll length                   & \multicolumn{1}{c|}{$1024$}      & \multicolumn{1}{c|}{$5$}    & $10$     & \multicolumn{2}{c|}{}                                         \\ \cline{1-4}
num minibatches                 & \multicolumn{1}{c|}{-}           & \multicolumn{2}{c|}{$32$}              & \multicolumn{2}{c|}{}                                         \\ \cline{1-4}
epochs                          & \multicolumn{2}{c|}{$4$}                                       & $8$      & \multicolumn{2}{c|}{}                                         \\ \hline
\end{tabular}
    \caption{
    PPO baseline parameters for each environment. 
    Due to differences in the implementation of stable baselines 3 JAX and Google Brax PPO, there is no hyperparameter for the number of minibatches in the case of LunarLander.
    The search space for the batch size is always a categorical choice over the power of two below and above the baseline value.
    }
    \label{tab:ppo_baselines}
\end{table}

\begin{table}[tb]
    \centering
    \footnotesize
\begin{tabular}{|l|cccc|cc|}
\hline
\multirow{2}{*}{Hyperparameter} & \multicolumn{4}{c|}{Baseline Values}                                                                          & \multicolumn{2}{c|}{Search Space}                             \\ \cline{2-7} 
                                & \multicolumn{1}{c|}{LunarLander} & \multicolumn{1}{c|}{Ant}       & \multicolumn{1}{c|}{Humanoid} & Wipe      & \multicolumn{1}{c|}{Range}             & Log-Scale            \\ \hline\hline
learning rate                   & \multicolumn{4}{c|}{$3\cdot10^{-4}$}                                                                          & \multicolumn{1}{c|}{$[1e^{-6}, 0.01]$} & \multirow{3}{*}{Yes} \\ \cline{1-6}
discounting                     & \multicolumn{1}{c|}{$0.99$}      & \multicolumn{1}{c|}{$0.95$}    & \multicolumn{2}{c|}{$0.99$}               & \multicolumn{1}{c|}{$[0.001, 0.02]$}   &                      \\ \cline{1-6}
tau                             & \multicolumn{1}{c|}{$0.01$}      & \multicolumn{1}{c|}{$0.005$}   & \multicolumn{2}{c|}{$0.005$}              & \multicolumn{1}{c|}{$[0.001, 0.1]$}    &                      \\ \hline
batch size                      & \multicolumn{1}{c|}{$256$}       & \multicolumn{1}{c|}{$512$}     & \multicolumn{1}{c|}{$1024$}   & $256$     & \multicolumn{1}{c|}{$\{b/2, b, 2b\}$}  & No                   \\ \hline
training steps                  & \multicolumn{1}{c|}{$500000$}    & \multicolumn{1}{c|}{$5242880$} & \multicolumn{1}{c|}{5242880}  & $1250000$ & \multicolumn{2}{c|}{\multirow{7}{*}{-}}                       \\ \cline{1-5}
episode length                  & \multicolumn{3}{c|}{$1000$}                                                                       & $500$     & \multicolumn{2}{c|}{}                                         \\ \cline{1-5}
num envs                        & \multicolumn{1}{c|}{$1$}         & \multicolumn{1}{c|}{$256$}     & \multicolumn{1}{c|}{$64$}     & $5$       & \multicolumn{2}{c|}{}                                         \\ \cline{1-5}
training freq                   & \multicolumn{4}{c|}{$1$}                                                                                      & \multicolumn{2}{c|}{}                                         \\ \cline{1-5}
gradient steps                  & \multicolumn{1}{c|}{$1$}         & \multicolumn{1}{c|}{$64$}      & \multicolumn{1}{c|}{$8$}      & $2$       & \multicolumn{2}{c|}{}                                         \\ \cline{1-5}
min replay buffer size          & \multicolumn{1}{c|}{$10000$}     & \multicolumn{2}{c|}{$8192$}                                    & $100$     & \multicolumn{2}{c|}{}                                         \\ \cline{1-5}
max replay buffer size          & \multicolumn{1}{c|}{$10^6$}      & \multicolumn{2}{c|}{$1038576$}                                 & $10^6$    & \multicolumn{2}{c|}{}                                         \\ \hline
\end{tabular}
    \caption{
    SAC baseline parameters for each environment. 
    The search space for the batch size is always a categorical choice over the power of two below and above the baseline value.
    }
    \label{tab:sac_baselines}
\end{table}

\section{Additional Optimisation Results and Visualisations}

In the following sections, we present additional plots and tables on the performance distributions of our different optimisation experiments.

\subsection{Default Shaped Reward Function Evaluation}
\label{app:default_shape_results}

In Table \ref{tab:handtuned_objective}, we present the returns of the obtained policies for each optimisation experiment, evaluated on the corresponding environment's default-shaped reward function.
\begin{table}[tb]
    \centering
    \footnotesize
\begin{tabular}{|c|cc|cccc|}
\hline
\multirow{3}{*}{Environment}                                                                  & \multicolumn{1}{c|}{\multirow{3}{*}{HPO}} & \multirow{3}{*}{RPO} & \multicolumn{4}{c|}{Default Shaped Reward Return}                                                                                                                                                                                                                                                                                                                                                                    \\ \cline{4-7} 
                                                                                              & \multicolumn{1}{c|}{}                     &                      & \multicolumn{2}{c|}{PPO}                                                                                                                                                                                       & \multicolumn{2}{c|}{SAC}                                                                                                                                                                            \\ \cline{4-7} 
                                                                                              & \multicolumn{1}{c|}{}                     &                      & \multicolumn{1}{c|}{Single Obj.}                                                                       & \multicolumn{1}{c|}{Multi Obj.}                                                                       & \multicolumn{1}{c|}{Single Obj.}                                                                                & Multi Obj.                                                                        \\ \hline\hline
\multirow{5}{*}{\begin{tabular}[c]{@{}c@{}}Gymnasium\\ LunarLander\\ (maximise)\end{tabular}} & \multicolumn{2}{c|}{base}                                        & \multicolumn{2}{c|}{274 (8\%)}                                                                                                                                                                               & \multicolumn{2}{c|}{276 (7\%)}                                                                                                                                                                    \\ \cline{2-7} 
                                                                                              & \multicolumn{1}{c|}{base}                 & DEHB                 & \multicolumn{1}{c|}{\begin{tabular}[c]{@{}c@{}}266 (10\%)\\ ($p=2.3\cdot10^{-04}$)\end{tabular}}   & \multicolumn{1}{c|}{\textbf{283 (7\%)}}                                                           & \multicolumn{1}{c|}{\begin{tabular}[c]{@{}c@{}}286 (7\%)\\ ($p=3.4\cdot10^{-02}$)\end{tabular}}             & \textbf{287 (7\%)}                                                            \\ \cline{2-7} 
                                                                                              & \multicolumn{1}{c|}{DEHB}                 & base                 & \multicolumn{1}{c|}{\begin{tabular}[c]{@{}c@{}}275 (7\%)\\ ($p=5.7\cdot10^{-04}$)\end{tabular}}    & \multicolumn{1}{c|}{\begin{tabular}[c]{@{}c@{}}272 (7\%)\\ ($p=3.1\cdot10^{-04}$)\end{tabular}}   & \multicolumn{1}{c|}{\begin{tabular}[c]{@{}c@{}}280 (7\%)\\ ($p=1.0\cdot10^{-20}$)\end{tabular}}             & \begin{tabular}[c]{@{}c@{}}283 (7\%)\\ ($p=5.2\cdot10^{-09}$)\end{tabular}    \\ \cline{2-7} 
                                                                                              & \multicolumn{1}{c|}{DEHB}                 & RS                   & \multicolumn{1}{c|}{\begin{tabular}[c]{@{}c@{}}275 (8\%)\\ ($p=1.7\cdot10^{-02}$)\end{tabular}}    & \multicolumn{1}{c|}{\begin{tabular}[c]{@{}c@{}}276 (8\%)\\ ($p=1.7\cdot10^{-03}$)\end{tabular}}   & \multicolumn{1}{c|}{\textbf{287 (7\%)}}                                                                     & \begin{tabular}[c]{@{}c@{}}283 (7\%)\\ ($p=7.0\cdot10^{-07}$)\end{tabular}    \\ \cline{2-7} 
                                                                                              & \multicolumn{2}{c|}{DEHB (ours)}                                 & \multicolumn{1}{c|}{\textbf{234 (25\%)}}                                                           & \multicolumn{1}{c|}{\textbf{283 (7\%)}}                                                           & \multicolumn{1}{c|}{\textbf{287 (7\%)}}                                                                     & \begin{tabular}[c]{@{}c@{}}285 (7\%)\\ ($p=2.2\cdot10^{-02}$)\end{tabular}    \\ \hline\hline
\multirow{5}{*}{\begin{tabular}[c]{@{}c@{}}Google Brax \\ Ant\\ (maximise)\end{tabular}}      & \multicolumn{2}{c|}{base}                                        & \multicolumn{2}{c|}{7293 (17\%)}                                                                                                                                                                             & \multicolumn{2}{c|}{8065 (30\%)}                                                                                                                                                                  \\ \cline{2-7} 
                                                                                              & \multicolumn{1}{c|}{base}                 & DEHB                 & \multicolumn{1}{c|}{\begin{tabular}[c]{@{}c@{}}7235 (18\%)\\ ($p=3.1\cdot10^{-114}$)\end{tabular}} & \multicolumn{1}{c|}{\begin{tabular}[c]{@{}c@{}}7236 (17\%)\\ ($p=3.4\cdot10^{-52}$)\end{tabular}} & \multicolumn{1}{c|}{\begin{tabular}[c]{@{}c@{}}7600 (34\%)\\ ($p=5.9\cdot10^{-19}$)\end{tabular}}           & \begin{tabular}[c]{@{}c@{}}7814 (32\%)\\ ($p=1.0\cdot10^{-03}$)\end{tabular}  \\ \cline{2-7} 
                                                                                              & \multicolumn{1}{c|}{DEHB}                 & base                 & \multicolumn{1}{c|}{\textbf{8379 (17\%)}}                                                          & \multicolumn{1}{c|}{\textbf{8127 (10\%)}}                                                         & \multicolumn{1}{c|}{\textbf{8169 (25\%)}}                                                                   & \textbf{8037 (19\%)}                                                          \\ \cline{2-7} 
                                                                                              & \multicolumn{1}{c|}{DEHB}                 & RS                   & \multicolumn{1}{c|}{\begin{tabular}[c]{@{}c@{}}8063 (18\%)\\ ($p=3.6\cdot10^{-09}$)\end{tabular}}  & \multicolumn{1}{c|}{-}                                                                                & \multicolumn{1}{c|}{\begin{tabular}[c]{@{}c@{}}7636 (24\%)\\ ($p=2.2\cdot10^{-10}$)\end{tabular}}           & -                                                                                 \\ \cline{2-7} 
                                                                                              & \multicolumn{2}{c|}{DEHB (ours)}                                        & \multicolumn{1}{c|}{\begin{tabular}[c]{@{}c@{}}8254 (16\%)\\ ($p=4.0\cdot10^{-03}$)\end{tabular}}  & \multicolumn{1}{c|}{\begin{tabular}[c]{@{}c@{}}8124 (9\%)\\ ($p=3.9\cdot10^{-02}$)\end{tabular}}  & \multicolumn{1}{c|}{\begin{tabular}[c]{@{}c@{}}7717 (28\%)\\ ($p=1.0\cdot10^{-09}$)\end{tabular}}           & \begin{tabular}[c]{@{}c@{}}7866 (19\%)\\ ($p=2.1\cdot10^{-03}$)\end{tabular}  \\ \hline\hline
\multirow{5}{*}{\begin{tabular}[c]{@{}c@{}}Google Brax \\ Humanoid\\ (maximise)\end{tabular}} & \multicolumn{2}{c|}{base}                                        & \multicolumn{2}{c|}{10016 (<1\%)}                                                                                                                                                                             & \multicolumn{2}{c|}{3273 (11\%)}                                                                                                                                                                \\ \cline{2-7} 
                                                                                              & \multicolumn{1}{c|}{base}                 & DEHB                 & \multicolumn{1}{c|}{\begin{tabular}[c]{@{}c@{}}10256 (<1\%)\\ ($p=6.6\cdot10^{-24}$)\end{tabular}}  & \multicolumn{1}{c|}{\begin{tabular}[c]{@{}c@{}}10439 (<1\%)\\ ($p=5.9\cdot10^{-23}$)\end{tabular}} & \multicolumn{1}{c|}{\begin{tabular}[c]{@{}c@{}}10509 (11\%)\\ ($p=2.3\cdot10^{-07}$)\end{tabular}}          & \begin{tabular}[c]{@{}c@{}}10695 (11\%)\\ ($p=6.4\cdot10^{-13}$)\end{tabular} \\ \cline{2-7} 
                                                                                              & \multicolumn{1}{c|}{DEHB}                 & base                 & \multicolumn{1}{c|}{\begin{tabular}[c]{@{}c@{}}10850 (2\%)\\ ($p=1.1\cdot10^{-07}$)\end{tabular}}  & \multicolumn{1}{c|}{\begin{tabular}[c]{@{}c@{}}10726 (<1\%)\\ ($p=2.1\cdot10^{-22}$)\end{tabular}} & \multicolumn{1}{c|}{\begin{tabular}[c]{@{}c@{}}10317 (19\%)\\ ($p=7.5\cdot10^{-17}$)\end{tabular}}          & \begin{tabular}[c]{@{}c@{}}9763 (17\%)\\ ($p=2.6\cdot10^{-64}$)\end{tabular}  \\ \cline{2-7} 
                                                                                              & \multicolumn{1}{c|}{DEHB}                 & RS                   & \multicolumn{1}{c|}{\begin{tabular}[c]{@{}c@{}}11204 (<1\%)\\ ($p=6.3\cdot10^{-03}$)\end{tabular}}  & \multicolumn{1}{c|}{-}                                                                                & \multicolumn{1}{c|}{\begin{tabular}[c]{@{}c@{}}\textbf{11746 (17\%)}\\ ($p=4.3\cdot10^{-01}$)\end{tabular}} & -                                                                                 \\ \cline{2-7} 
                                                                                              & \multicolumn{2}{c|}{DEHB (ours)}                                 & \multicolumn{1}{c|}{\textbf{11599 (7\%)}}                                                          & \multicolumn{1}{c|}{\textbf{11562 (<1\%)}}                                                         & \multicolumn{1}{c|}{\textbf{12141 (11\%)}}                                                                  & \textbf{12292 (8\%)}                                                          \\ \hline\hline
\multirow{5}{*}{\begin{tabular}[c]{@{}c@{}}Robosuite \\ Wipe\\ (maximise)\end{tabular}}          & \multicolumn{2}{c|}{base}                                        & \multicolumn{2}{c|}{\multirow{5}{*}{-}}                                                                                                                                                                        & \multicolumn{2}{c|}{108 (38\%)}                                                                                                                                       \\ \cline{2-3} \cline{6-7}
                                                                                              & \multicolumn{1}{c|}{base}                 & DEHB                 & \multicolumn{2}{c|}{}                                                                                                                                                                                          & \multicolumn{1}{c|}{\begin{tabular}[c]{@{}c@{}}77 (57\%)\\ ($p=1.7\cdot10^{-98}$)\end{tabular}}             & \begin{tabular}[c]{@{}c@{}}78 (58\%)\\ ($p=1.3\cdot10^{-159}$)\end{tabular}                                                                                 \\ \cline{2-3} \cline{6-7}
                                                                                              & \multicolumn{1}{c|}{DEHB}                 & base                 & \multicolumn{2}{c|}{}                                                                                                                                                                                          & \multicolumn{1}{c|}{\begin{tabular}[c]{@{}c@{}}\textbf{134 (20\%)}\end{tabular}}                            & \begin{tabular}[c]{@{}c@{}}\textbf{131 (20\%)}\\ ($p=9.7\cdot10^{-1}$)\end{tabular}                                                                                                \\ \cline{2-3} \cline{6-7}
                                                                                              & \multicolumn{1}{c|}{DEHB}                 & RS                   & \multicolumn{2}{c|}{}                                                                                                                                                                                          & \multicolumn{1}{c|}{\begin{tabular}[c]{@{}c@{}}126 (21\%)\\ ($p=4.7\cdot10^{-11}$)\end{tabular}}            & -                                                             \\ \cline{2-3} \cline{6-7}
                                                                                              & \multicolumn{2}{c|}{DEHB (ours)}                                 & \multicolumn{2}{c|}{}                                                                                                                                                                                          & \multicolumn{1}{c|}{\begin{tabular}[c]{@{}c@{}}127 (25\%)\\ ($p=2.3\cdot10^{-23}$)\end{tabular}}            & \begin{tabular}[c]{@{}c@{}}\textbf{132 (20\%)}\end{tabular}                                                                                \\ \hline
\end{tabular}

    \caption{
        Results of the trained policies for each optimisation experiment evaluated on each environment's default shaped reward function and the coefficients of variations in parenthesis. 
        Columns \textit{HPO} and \textit{RPO} indicate the respective optimisation methods: \textit{base} for baseline values, \textit{DEHB} and \textit{RS} for optimisation with DEHB or random search. 
        Hence, the first row of each environment is the baseline performance, followed by rows optimising reward parameters, hyperparameters, or both. 
        Each experiment's performance is computed similarly as in Table \ref{tab:external_objective}.
        Performances without significant statistical differences to the best-performing optimisation experiment, are highlighted in bold for each environment.
        We reported the test's p-values of the comparison in each cell.
    }
    \label{tab:handtuned_objective}
\end{table}
Our observations indicate that for LunarLander and Humanoid, the combined optimisation consistently matches or outperforms the best performance, except for multi-objective SAC training for LunarLander. 
This suggests that the benefits on the task performance effectively transfer to the default shaped reward function, even though the policies were not specifically optimised for it.

In the case of Ant, the performance is slightly lower than that achieved through hyperparameter-only optimisation, yet qualitative analysis shows that the environment is still clearly solved. 
For the Robosuite Wipe environment, however, the combined optimisation performs significantly worse than the hyperparameter-only optimisation, which starkly contrasts with the evaluation of the environment's task performance.

Further analysis reveals that this discrepancy is due to the default shaped reward function's inadequate representation of the environment's task objective. Specifically, policies that do not completely clean the table but maintain contact with it until the end of an episode can accumulate a higher overall return compared to those that quickly complete the cleaning task. 
Consequently, the policies resulting from combined optimisation, which prioritise rapid table cleaning, achieve lower returns despite better performance in wiping the table.

\subsection{Optimisation Performance Boxplots}
\label{app:boxplots}

\begin{figure}[H]
    \centering
    \includegraphics[width=\textwidth]{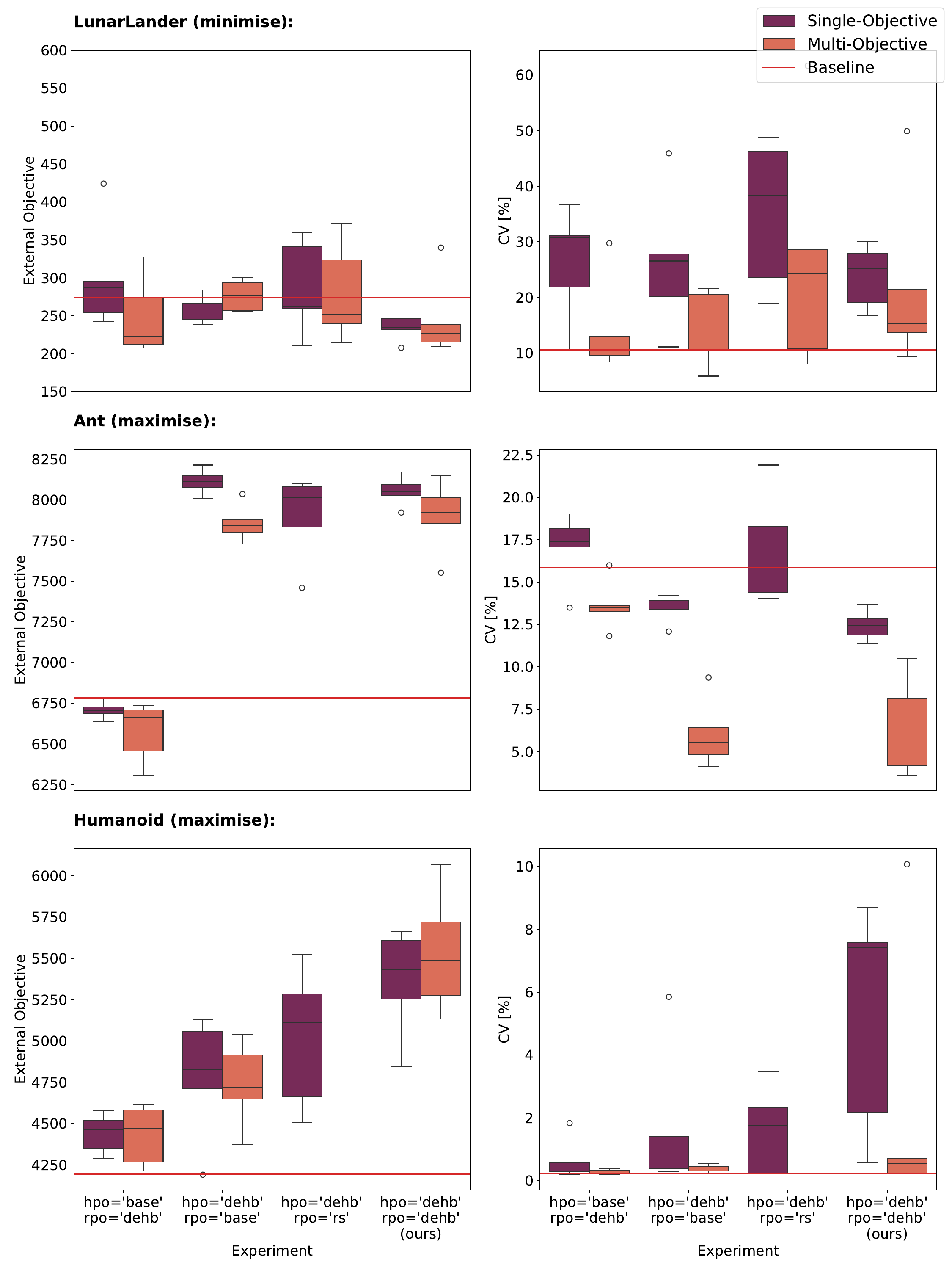}
    \caption{
        Boxplots for the PPO optimisation of the five median performances of each experiment's optimisation runs.
        We denote by DEHB, rs and base if hyperparameters (hpo) or reward parameters (rpo) were optimised with DEHB, random search or fixed to their baseline values.
        The red line denotes the baseline performance.
    }
    \label{fig:ppo_sep_boxplot}
\end{figure}

\begin{figure}[H]
    \centering
    \includegraphics[width=\textwidth]{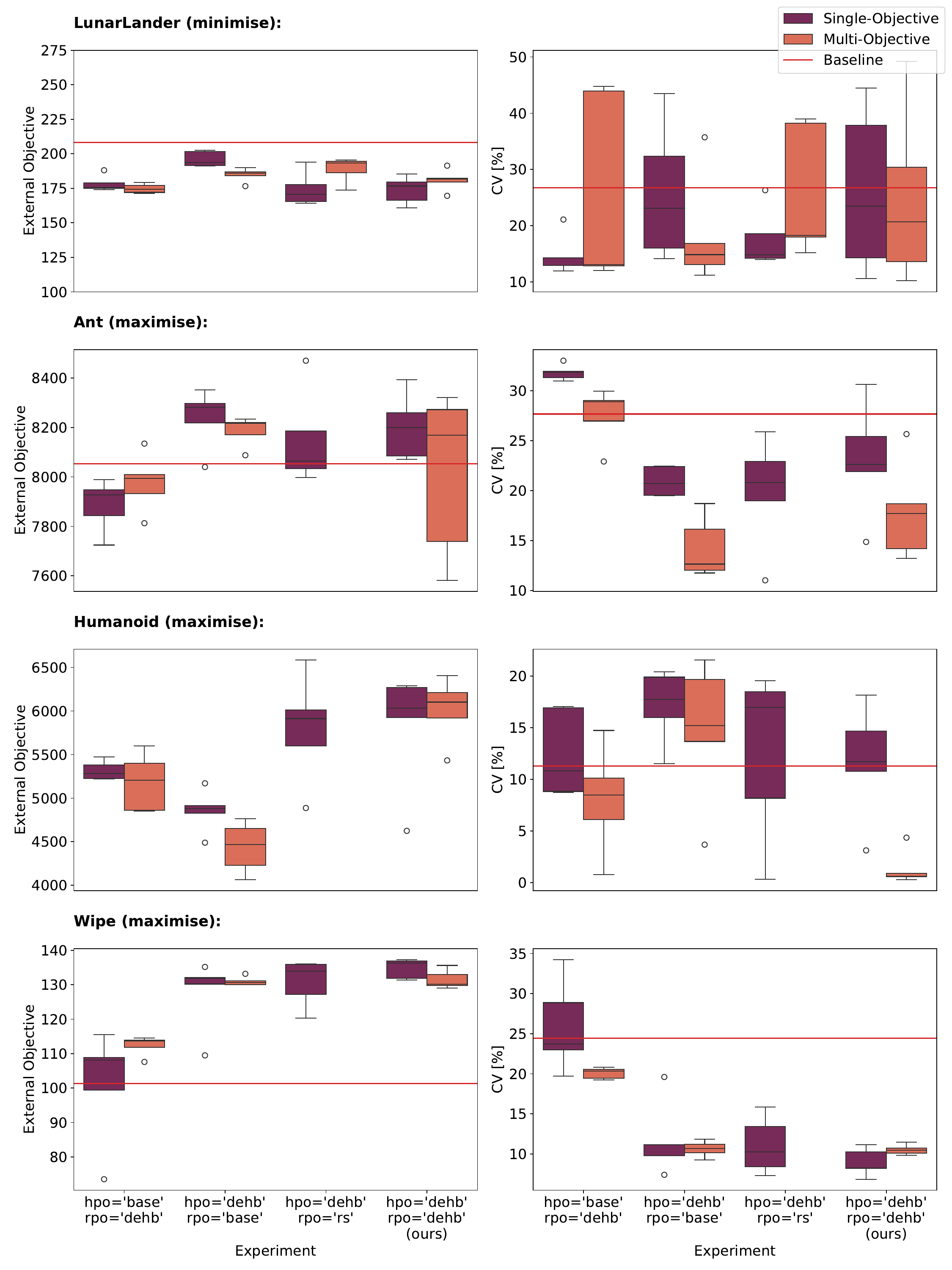}
    \caption{
        Boxplots for the SAC optimisation of the five median performances of each experiment's optimisation runs.
        We denote by DEHB, rs and base if hyperparameters (hpo) or reward parameters (rpo) were optimised with DEHB, random search or fixed to their baseline values.
        The red line denotes the baseline performance.
    }
    \label{fig:sac_sep_boxplot}
\end{figure}

\begin{figure}[H]
    \centering
    \includegraphics[width=\textwidth]{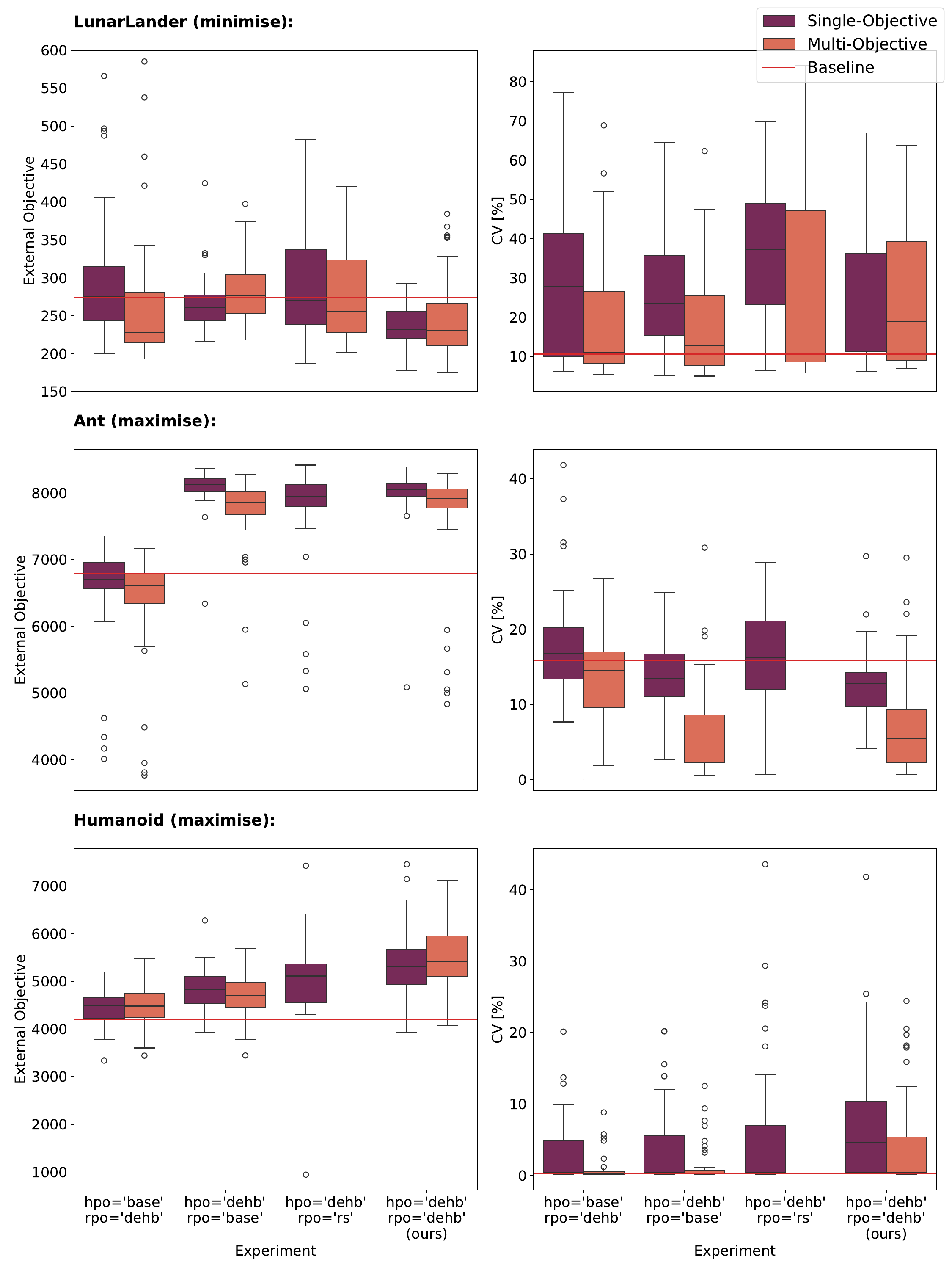}
    \caption{
        Boxplots for the PPO optimisation experiments over all 50 evaluation trainings of each experiment.
        We denote by DEHB, rs and base if hyperparameters (hpo) or reward parameters (rpo) were optimised with DEHB, random search or fixed to their baseline values.
        The red line denotes the baseline performance.
    }
    \label{fig:ppo_pooled_boxplot}
\end{figure}

\begin{figure}[H]
    \centering
    \includegraphics[width=\textwidth]{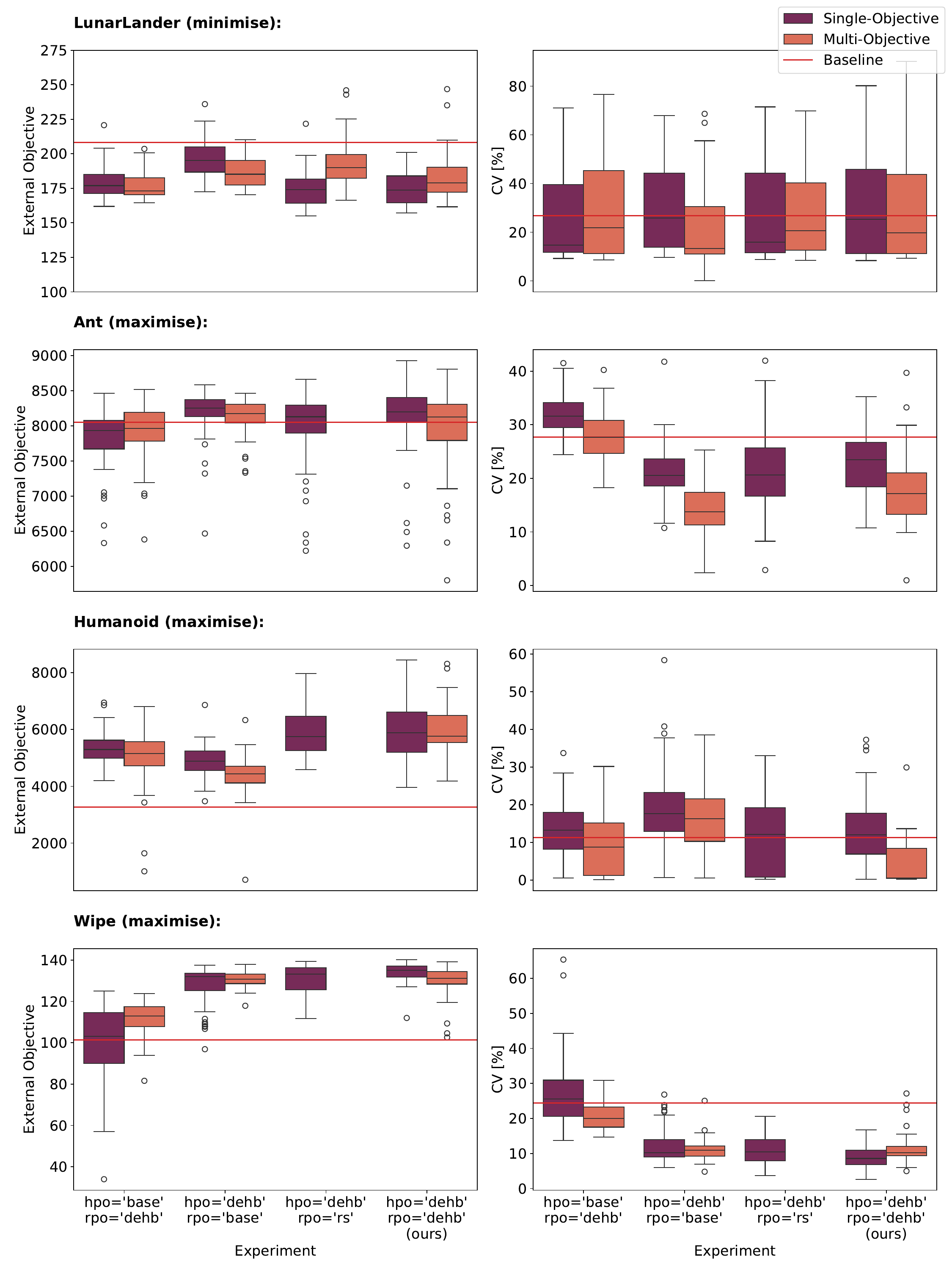}
    \caption{
        Boxplots for the SAC optimisation experiments over all 50 evaluation trainings of each experiment.
        We denote by DEHB, rs and base if hyperparameters (hpo) or reward parameters (rpo) were optimised with DEHB, random search or fixed to their baseline values.
        The red line denotes the baseline performance.
    }
    \label{fig:sac_pooled_boxplot}
\end{figure}

In addition to the results presented in Table \ref{tab:external_objective}, we provide an overview of the full dataset as boxplots. 
Figures \ref{fig:ppo_sep_boxplot} and \ref{fig:sac_sep_boxplot} display boxplots for the median performances of each experiment's five optimisation runs. 
For each experiment's optimisation run, we calculate the median performance across its ten evaluation trainings and present boxplots for the resulting five values per experiment. 
Figures \ref{fig:ppo_pooled_boxplot} and \ref{fig:sac_pooled_boxplot} showcase boxplots for the combined 50 evaluation training performances, obtained by aggregating all ten evaluation training performances for each of the five optimisation runs per experiment. 
In each boxplot, the baseline's median performance is marked with a red line.

Consistent with our analysis in Section \ref{sec:optimisation}, we observe that combined optimisation can match or surpass the individual optimisations of hyperparameters and reward parameters. Furthermore, multi-objective optimisation substantially enhances stability with minimal or no reductions in performance.

\subsection{Incumbent Performance during Optimisation}
\label{all:incumbents}

\begin{figure}[tb]
    \centering
    \includegraphics[width=\textwidth]{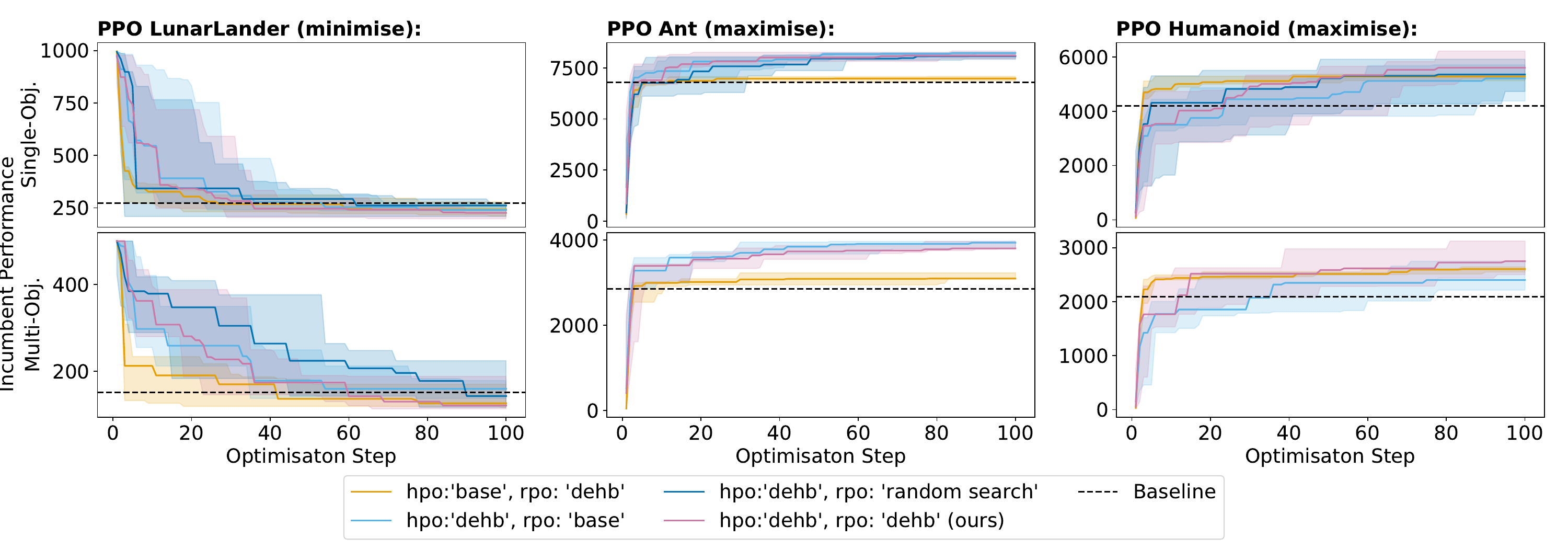}
    \caption{
    The median optimisation objective's incumbent performance across the five optimisation runs for the PPO experiments at each time step. 
    The min and max are given as error bars. 
    }
    \label{fig:ppo_incumb}
\end{figure}

Figure \ref{fig:ppo_incumb} depicts the median incumbent performance during each PPO optimisation experiment. 
We note that the optimisation steps necessary to surpass the baselines generally occur much earlier than the complete duration of the optimisation, similar to the findings of \citet{eimer23}. 
However, we also observe that continuous improvement is still achieved after exceeding the baseline performance. 
Moreover, the combined optimisation does not seem to be significantly slower than optimising hyperparameters or reward parameters alone, suggesting that the combined optimisation can enhance results without additional costs, similar to the SAC results presented in Figure \ref{fig:incumb}.

\newpage

\section{Reward Scaling}
\label{app:scaling}

We examined two alternative methods to perform reward scaling for the Google Brax experiments denoted as explicit and implicit scaling:

\textit{Explicit reward scaling} aims to disentangle effects between the chosen scaling $\alpha$ and the reward weights $w$ during the optimisation by normalising the reward weights.
Formally, the optimiser's selected weights $w$ are normalised by 
\begin{equation}
w' = \Vert\hat{w}\Vert_1 \cdot \frac{w}{\left\|w\right\|_1},
\end{equation}
where $\hat{w}$ are the default reward weights of the given environment.
The resulting reward function $\tilde{r}^{\alpha, w'}$ was then used for RL training as described in Section \ref{sec:joint_opt_perf}.
In contrast, \textit{implicit reward scaling} is done by keeping the reward scale fixed as $\alpha = 1$ and instead optimising reward parameters with search spaces multiplied by the upper bound of the explicit scaling. 
The detailed search spaces are given in Table \ref{tab:exp_vs_impl}.
\begin{table}[tb]
    \centering
    \footnotesize
\begin{tabular}{|l|cc|cc|}
\hline
\multirow{2}{*}{Reward Weight} & \multicolumn{2}{c|}{Explicit Scaling} & \multicolumn{2}{c|}{Implicit Scaling}         \\ \cline{2-5} 
                               & \multicolumn{1}{c|}{Ant}  & Humanoid  & \multicolumn{1}{c|}{Ant}        & Humanoid    \\ \hline\hline
$w_{dist}$                     & \multicolumn{2}{c|}{$[0, 10]$}        & \multicolumn{1}{c|}{$[0, 25]$}  & $[0, 63.5]$ \\ \hline
$w_{healthy}$                  & \multicolumn{2}{c|}{$[0, 10]$}        & \multicolumn{1}{c|}{$[0, 25]$}  & $[0, 63.5]$ \\ \hline
$w_{force}$                    & \multicolumn{2}{c|}{$[0, 1]$}         & \multicolumn{1}{c|}{$[0, 2.5]$} & $[0, 6.35]$ \\ \hline
$\alpha$                       & \multicolumn{2}{c|}{$[0, 10]$}        & \multicolumn{1}{c|}{$1$}        & $1$         \\ \hline
\end{tabular}
    \caption{
    The different search spaces of explicit and implicit reward scaling.
    In the case of explicit reward scaling, the reward weights are normalised to one and scaled by $\Vert\hat{w}_{ant}\Vert_1 = 2.5$ and $\Vert\hat{w}_{humanoid}\Vert_1 = 6.35$ before applying the reward scale $\alpha$ for Ant and Humanoid, respectively. 
    Thereby, the search spaces in the explicit and implicit scaling have the same upper and lower bounds.
    }
    \label{tab:exp_vs_impl}
\end{table}

In Table \ref{tab:rs_external_objective}, we report the results of the two scaling approaches. 
\begin{table}[tb]
    \centering
    \footnotesize
\begin{tabular}{|c|cc|cc|}
\hline
\multirow{2}{*}{Environment}                                                               & \multicolumn{1}{c|}{\multirow{2}{*}{HPO}} & \multirow{2}{*}{RPO} & \multicolumn{2}{c|}{Task Performance}                                                                                                   \\ \cline{4-5} 
                                                                                & \multicolumn{1}{c|}{}                     &                      & \multicolumn{1}{c|}{Implicit Scaling}                                                                          & Explicit Scaling         \\ \hline\hline
\multirow{2}{*}{\begin{tabular}[c]{@{}c@{}}Google Brax\\ Ant \\ (maximise)\end{tabular}}      & \multicolumn{1}{c|}{base}                 & DEHB                 & \multicolumn{1}{c|}{\begin{tabular}[c]{@{}c@{}}7741 (28\%)\\ ($p=2.8\cdot10^{-03}$)\end{tabular}}          & \textbf{7927 (32\%)} \\ \cline{2-5} 
                                                                                & \multicolumn{2}{c|}{DEHB}                                        & \multicolumn{1}{c|}{\begin{tabular}[c]{@{}c@{}}7625 (23\%)\\ ($p=1.0\cdot10^{-12}$)\end{tabular}}          & \textbf{8199 (23\%)} \\ \hline
\multirow{2}{*}{\begin{tabular}[c]{@{}c@{}}Google Brax\\ Humanoid \\ (maximise)\end{tabular}} & \multicolumn{1}{c|}{base}                 & DEHB                 & \multicolumn{1}{c|}{\begin{tabular}[c]{@{}c@{}}\textbf{5124 (12\%)}\\ ($p=4.0\cdot10^{-01}$)\end{tabular}} & \textbf{5284 (11\%)} \\ \cline{2-5} 
                                                                                & \multicolumn{2}{c|}{DEHB}                                        & \multicolumn{1}{c|}{\begin{tabular}[c]{@{}c@{}}\textbf{5846 (12\%)}\\ ($p=8.8\cdot10^{-01}$)\end{tabular}} & \textbf{6033 (12\%)} \\ \hline
\end{tabular}
    \caption{
    Results for the different single-objective optimisation experiments with explicit and implicit reward scaling. 
    Performances in each row without significant statistical differences to the best scaling experiment, as determined by a linear mixed-effects model analysis, are highlighted in bold.
    We reported the test's p-values of the comparison in each cell.
    }
    \label{tab:rs_external_objective}
\end{table}
Explicit scaling statistically significantly outperforms implicit scaling for Ant, whereas, for Humanoid, the median performance is better but not significant.
Overall, this hints to explicit scaling being a better choice in cases where scaling matters to the algorithm.
Due to its better performance, we used the explicit scaling method for performing experiments in Section \ref{sec:joint_opt_perf}.

\section{Best performing Hyperparameter and Reward Weight Configurations per Algorithm and Environment}
\label{app:best_configurations}

We present the best-performing configurations we found for each training algorithm and environment based on median external objective performance across ten evaluation trainings.
Table \ref{tab:ppo_best_params} and Table \ref{tab:sac_best_params} display the parameters for PPO and SAC, respectively, and additionally report the performance on the external objective and default shaped reward.
We hope the configurations can help as baselines and for future research.
The hyperparameter configuration and reward parameters must be used together in training for each environment to achieve the best performance.
\begin{table}[tb]
    \centering
    \begin{subtable}{\textwidth}
        \centering
\footnotesize
\begin{tabular}{|lccc|}
\hline
\multicolumn{1}{|l|}{Hyperparameter}       & \multicolumn{1}{c|}{LunarLander}        & \multicolumn{1}{c|}{Ant}                & Humanoid           \\ \hline\hline
\multicolumn{1}{|l|}{learning rate}        & \multicolumn{1}{c|}{$3e^{-4}$}          & \multicolumn{1}{c|}{0.00112}            & 0.000359           \\ \hline
\multicolumn{1}{|l|}{discounting}          & \multicolumn{1}{c|}{0.999}              & \multicolumn{1}{c|}{0.964}              & 0.962              \\ \hline
\multicolumn{1}{|l|}{gae lambda}                  & \multicolumn{1}{c|}{0.98}               & \multicolumn{1}{c|}{0.8378}             & 0.966              \\ \hline
\multicolumn{1}{|l|}{clipping rate}        & \multicolumn{1}{c|}{0.2}                & \multicolumn{1}{c|}{0.276}              & 0.156              \\ \hline
\multicolumn{1}{|l|}{entropy coef}         & \multicolumn{1}{c|}{0.01}               & \multicolumn{1}{c|}{0.345}              & 0.00657            \\ \hline
\multicolumn{1}{|l|}{value coef}           & \multicolumn{1}{c|}{0.5}                & \multicolumn{1}{c|}{0.469}              & 0.35               \\ \hline
\multicolumn{1}{|l|}{batch size}           & \multicolumn{1}{c|}{64}                 & \multicolumn{1}{c|}{1024}               & 512                \\ \hline\hline
Reward Parameter                              & \multicolumn{3}{l|}{}                                                                                  \\ \hline\hline
\multicolumn{1}{|l|}{$w_{dist}$}           & \multicolumn{1}{c|}{3.078}              & \multicolumn{1}{c|}{1.0}                & 9.28               \\ \hline
\multicolumn{1}{|l|}{$w_{vel}$}            & \multicolumn{1}{c|}{0.989}              & \multicolumn{1}{c|}{\multirow{3}{*}{-}} & \multirow{3}{*}{-} \\ \cline{1-2}
\multicolumn{1}{|l|}{$w_{tilting}$}        & \multicolumn{1}{c|}{0.222}              & \multicolumn{1}{c|}{}                   &                    \\ \cline{1-2}
\multicolumn{1}{|l|}{$w_{leg}$}            & \multicolumn{1}{c|}{4.53}               & \multicolumn{1}{c|}{}                   &                    \\ \hline
\multicolumn{1}{|l|}{$w_{healthy}$}        & \multicolumn{1}{c|}{\multirow{2}{*}{-}} & \multicolumn{1}{c|}{1.0}                & 2.82               \\ \cline{1-1} \cline{3-4} 
\multicolumn{1}{|l|}{$w_{force}$}          & \multicolumn{1}{c|}{}                   & \multicolumn{1}{c|}{0.5}                & 0.938              \\ \hline
\multicolumn{1}{|l|}{$\alpha$} & \multicolumn{1}{c|}{1.0}                & \multicolumn{1}{c|}{1.0} & 1.0                \\ \hline\hline
Performance  & \multicolumn{3}{l|}{} \\
\hline\hline
\multicolumn{1}{|l|}{external objective}           & \multicolumn{1}{c|}{207 (4\%)}              & \multicolumn{1}{c|}{8213 (0.1\%)}                &  6068 (0.001\%)             \\ \hline
\multicolumn{1}{|l|}{default shaped return}           & \multicolumn{1}{c|}{286 (2\%)}              & \multicolumn{1}{c|}{8378 (0.2\%)}                &  12440 (0.003\%)             \\ \hline
\end{tabular}

    \end{subtable}
    \caption{
    Best performing configurations obtained by our optimisations for PPO training selected by their external objective performance.
    For each configuration, we report the achieved performance measured by the external objective and the default-shaped reward function with the coefficients of variations in parenthesis.
    }
    \label{tab:ppo_best_params}
\end{table}

\begin{table}[tb]
    \centering
\footnotesize
\begin{tabular}{|lcccc|}
\hline
\multicolumn{1}{|l|}{Hyperparameter}  & \multicolumn{1}{c|}{LunarLander}        & \multicolumn{1}{c|}{Ant}                & \multicolumn{1}{c|}{Humanoid}           & Wipe               \\ \hline\hline
\multicolumn{1}{|l|}{learning rate}   & \multicolumn{1}{c|}{0.000998}           & \multicolumn{1}{c|}{0.000703}           & \multicolumn{1}{c|}{0.00078}            & 0.00031            \\ \hline
\multicolumn{1}{|l|}{discounting}     & \multicolumn{1}{c|}{0.988}              & \multicolumn{1}{c|}{0.982}              & \multicolumn{1}{c|}{0.968}              & 0.83               \\ \hline
\multicolumn{1}{|l|}{tau}             & \multicolumn{1}{c|}{0.0771}             & \multicolumn{1}{c|}{0.00444}            & \multicolumn{1}{c|}{0.0297}             & 0.00601            \\ \hline
\multicolumn{1}{|l|}{batch size}      & \multicolumn{1}{c|}{256}                & \multicolumn{1}{c|}{512}                & \multicolumn{1}{c|}{1024}               & 256                \\ \hline\hline
Reward Parameter                         & \multicolumn{4}{l|}{}                                                                                                                            \\ \hline\hline
\multicolumn{1}{|l|}{$w_{dist}$}      & \multicolumn{1}{c|}{5.901}              & \multicolumn{1}{c|}{2.253}              & \multicolumn{1}{c|}{7.61}               & 7.745              \\ \hline
\multicolumn{1}{|l|}{$w_{dist\_th}$}  & \multicolumn{1}{c|}{-}                  & \multicolumn{1}{c|}{\multirow{4}{*}{-}} & \multicolumn{1}{c|}{\multirow{4}{*}{-}} & 2.60               \\ \cline{1-2} \cline{5-5} 
\multicolumn{1}{|l|}{$w_{vel}$}       & \multicolumn{1}{c|}{2.691}              & \multicolumn{1}{c|}{}                   & \multicolumn{1}{c|}{}                   & 0.888              \\ \cline{1-2} \cline{5-5} 
\multicolumn{1}{|l|}{$w_{tilting}$}   & \multicolumn{1}{c|}{1.102}              & \multicolumn{1}{c|}{}                   & \multicolumn{1}{c|}{}                   & \multirow{3}{*}{-} \\ \cline{1-2}
\multicolumn{1}{|l|}{$w_{leg}$}       & \multicolumn{1}{c|}{5.465}              & \multicolumn{1}{c|}{}                   & \multicolumn{1}{c|}{}                   &                    \\ \cline{1-4}
\multicolumn{1}{|l|}{$w_{healthy}$}   & \multicolumn{1}{c|}{\multirow{5}{*}{-}} & \multicolumn{1}{c|}{0.136}              & \multicolumn{1}{c|}{2.99}               &                    \\ \cline{1-1} \cline{3-5} 
\multicolumn{1}{|l|}{$w_{force}$}     & \multicolumn{1}{c|}{}                   & \multicolumn{1}{c|}{0.109}              & \multicolumn{1}{c|}{0.916}              & 0.030              \\ \cline{1-1} \cline{3-5} 
\multicolumn{1}{|l|}{$w_{wiped}$}     & \multicolumn{1}{c|}{}                   & \multicolumn{1}{c|}{\multirow{3}{*}{-}} & \multicolumn{1}{c|}{\multirow{3}{*}{-}} & 82.3               \\ \cline{1-1} \cline{5-5} 
\multicolumn{1}{|l|}{$w_{contact}$}   & \multicolumn{1}{c|}{}                   & \multicolumn{1}{c|}{}                   & \multicolumn{1}{c|}{}                   & 0.594              \\ \cline{1-1} \cline{5-5} 
\multicolumn{1}{|l|}{$w_{collision}$} & \multicolumn{1}{c|}{}                   & \multicolumn{1}{c|}{}                   & \multicolumn{1}{c|}{}                   & -83.0              \\ \hline
\multicolumn{1}{|l|}{$\alpha$}        & \multicolumn{1}{c|}{1.0}                & \multicolumn{1}{c|}{7.44}               & \multicolumn{1}{c|}{7.40}               & 1.0                \\ \hline\hline
Performance & \multicolumn{4}{l|}{}                                                                                                                            \\ \hline\hline
\multicolumn{1}{|l|}{external objective}        & \multicolumn{1}{c|}{160 (6\%)}                & \multicolumn{1}{c|}{8469 (0.2\%)}               & \multicolumn{1}{c|}{6583 (0.1\%)}               & 137 (5\%)                \\ \hline
\multicolumn{1}{|l|}{default shaped return}        & \multicolumn{1}{c|}{288 (2\%)}                & \multicolumn{1}{c|}{7903 (0.3\%)}               & \multicolumn{1}{c|}{12670 (0.07\%)}               & 126 (14\%)                \\ \hline
\end{tabular}
    \caption{
    Best performing configurations obtained by our optimisations for SAC training selected by their external objective performance.
    For each configuration, we report the achieved performance measured by the external objective and the default-shaped reward function with the coefficients of variations in parenthesis.
    }
    \label{tab:sac_best_params}
\end{table}

\section{Execution Environment}
\label{app:exc_environment}

All experiments were conducted on a high-performance cluster running the Rocky Linux operating system, release 8.9. 
The Gymnasium LunarLander and Robosuite Wipe optimisations were executed on CPU nodes, while the Google Brax optimisations utilised GPU nodes.

The CPU-based optimisations were carried out on nodes equipped with Intel Xeon Platinum 8160 2.1 GHz processors, each equipped
with 24 cores each and 33\,792 KB cache, with approximately 3.75 GB of RAM per core. 
The GPU-based optimisations utilised NVIDIA Volta 100 GPUs (V100-SXM2) with 16 GB of memory.

During the optimisation process, the LunarLander and Wipe environments, on average, employed 25 and 32 CPU cores in parallel, respectively. 
For LunarLander, each RL training run used 4 cores, while Wipe training utilised 5 cores per run. 
During optimisation, the Google Brax environments required an average of 6 parallel GPUs, with each RL training run conducted on a single GPU.

\section{Linear Mixed Effects Regression Analysis}
\label{app:lme_test}

To thoroughly analyse the differences between our optimisation experiments, we employed a linear mixed effects regression with a Wald test to analyse the difference in performance between experiments.
We conducted the test based on the introduction of \citet{brown2021}, using the Wald test as a commonly used approximation of the likelihood-ratio test.
The linear mixed-effects model analysis enables us to compare the performance of two optimisation experiments across their respective 50 evaluation runs while accounting for the dependencies induced by the seed of an optimisation run to which an evaluation belongs.
Therefore, we can compare the full extent of our data and avoid collapsing it by summarising each optimisation run's performance by the median performance of its 10 evaluation trainings.

For each environment and algorithm, we always pick the best-performing optimisation experiment based on its median performance presented in Table \ref{tab:external_objective}.
We then compare this optimisation experiment pairwise to the other corresponding optimisation experiments and test whether the performance is statistically significantly different.
Hence, the value to be predicted by the linear mixed-effects model is evaluation performance, with the fixed effect being the two compared experiments.
The different evaluations are grouped by their corresponding optimisation seed as the model's random effect.
Using the Wald test, we then check if removing the fixed experiment effect from the model would substantially harm the prediction performance of the model.
Hence, small p-values of the test indicate that the model with the fixed experiment effect provides a better fit, and therefore, the experiments' performances are statistically significantly different. 
We applied a commonly used significance level of $0.05$ to test for significance.

For preprocessing the 100 evaluation data points of two experiments, we normalised mean performance to 0 and standard deviation to 1. 
Afterwards, we fit a mixed-effects model on the data and remove all points as outliers with residuals deviating more than two times the standard deviation from the mean. 
We then fit a model on the cleaned data and perform the Wald test to check for significance in the fixed effect, hence the difference between the two experiments.

The assumptions underlying the test are (a) independence of the random effects and (b) homoskedasticity of the residuals of the fitted linear mixed-effects model.
The normality of the residuals is an assumption of minor importance, as mixed-effect models have been shown to be robust to violations of this distributional assumption \citep{schielzeth2020};
\citet{gelman2006} even suggest not to test for normality of the residuals.
The independence of the optimisation runs as the random effect is ensured by using different random seeds.
To check the homoskedasticity assumption, we performed  White’s Lagrange multiplier test on the residuals.
The null hypothesis of White's test is homoskedasticity, and hence, large p-values suggest that the assumption of homoskedasticity is fulfilled.
Further, we can reasonably assume that the evaluations of our experiments follow a normal distribution, and we further tested normality of the residuals using the Shapiro-Wilk test. 

In Table \ref{tab:lme_stats}, we present the p-values of the Wald and White tests, as well as the number of outliers removed for the pairwise comparisons detailed in Table \ref{tab:external_objective}. 
\begin{table}[tb]
    \centering
    \footnotesize
\begin{tabular}{|c|cc|cccc|}
\hline
                                                                                               & \multicolumn{1}{c|}{}                      &                       & \multicolumn{4}{c|}{Test Statistics}                                                                                                                                                                                                                                                                                                                                                                                                                                                                                                                                                                                       \\ \cline{4-7} 
                                                                                               & \multicolumn{1}{c|}{}                      &                       & \multicolumn{2}{c|}{PPO}                                                                                                                                                                                                                                                                       & \multicolumn{2}{c|}{SAC}                                                                                                                                                                                                                                                                                                     \\ \cline{4-7} 
\multirow{-3}{*}{Environment}                                                                  & \multicolumn{1}{c|}{\multirow{-3}{*}{HPO}} & \multirow{-3}{*}{RPO} & \multicolumn{1}{c|}{Single Obj.}                                                                                     & \multicolumn{1}{c|}{Multi Obj.}                                                                                                                                         & \multicolumn{1}{c|}{Single Obj.}                                                                                                                                        & Multi Obj.                                                                                                                                         \\ \hline\hline
                                                                                               & \multicolumn{1}{c|}{base}                  & DEHB                  & \multicolumn{1}{c|}{\begin{tabular}[c]{@{}c@{}}$p=8.1\cdot10^{-06}$\\  $p_{hw}=0.006$\\  $|x_{out}|=4$\end{tabular}} & \multicolumn{1}{c|}{\textbf{best}}                                                                                                                                      & \multicolumn{1}{c|}{\cellcolor[HTML]{EFEFEF}\begin{tabular}[c]{@{}c@{}}\textbf{$p=1.6\cdot10^{-01}$}\\  \textbf{$p_{hw}=0.526$}\\  \textbf{$|x_{out}|=2$}\end{tabular}} & \textbf{best}                                                                                                                                      \\ \cline{2-7} 
                                                                                               & \multicolumn{1}{c|}{DEHB}                  & base                  & \multicolumn{1}{c|}{\begin{tabular}[c]{@{}c@{}}$p=4.5\cdot10^{-08}$\\  $p_{hw}=0.623$\\  $|x_{out}|=2$\end{tabular}} & \multicolumn{1}{c|}{\begin{tabular}[c]{@{}c@{}}$p=5.1\cdot10^{-06}$\\  $p_{hw}=0.485$\\  $|x_{out}|=5$\end{tabular}}                                                    & \multicolumn{1}{c|}{\begin{tabular}[c]{@{}c@{}}$p=3.4\cdot10^{-15}$\\  $p_{hw}=0.185$\\  $|x_{out}|=2$\end{tabular}}                                                    & \begin{tabular}[c]{@{}c@{}}$p=2.1\cdot10^{-05}$\\  $p_{hw}=0.498$\\  $|x_{out}|=2$\end{tabular}                                                    \\ \cline{2-7} 
                                                                                               & \multicolumn{1}{c|}{DEHB}                  & RS                    & \multicolumn{1}{c|}{\begin{tabular}[c]{@{}c@{}}$p=2.9\cdot10^{-07}$\\  $p_{hw}=0.106$\\  $|x_{out}|=5$\end{tabular}} & \multicolumn{1}{c|}{\begin{tabular}[c]{@{}c@{}}$p=8.0\cdot10^{-04}$\\  $p_{hw}=0.254$\\  $|x_{out}|=4$\end{tabular}}                                                    & \multicolumn{1}{c|}{\textbf{best}}                                                                                                                                      & \begin{tabular}[c]{@{}c@{}}$p=1.1\cdot10^{-08}$\\  $p_{hw}=0.276$\\  $|x_{out}|=5$\end{tabular}                                                    \\ \cline{2-7} 
\multirow{-4}[20]{*}{\begin{tabular}[c]{@{}c@{}}Gymnasium\\ LunarLander\\ (minimise)\end{tabular}} & \multicolumn{2}{c|}{DEHB (ours)}                                   & \multicolumn{1}{c|}{\textbf{best}}                                                                                   & \multicolumn{1}{c|}{\cellcolor[HTML]{EFEFEF}\begin{tabular}[c]{@{}c@{}}\textbf{$p=9.2\cdot10^{-01}$}\\  \textbf{$p_{hw}=0.994$}\\  \textbf{$|x_{out}|=4$}\end{tabular}} & \multicolumn{1}{c|}{\cellcolor[HTML]{EFEFEF}\begin{tabular}[c]{@{}c@{}}\textbf{$p=3.4\cdot10^{-01}$}\\  \textbf{$p_{hw}=0.612$}\\  \textbf{$|x_{out}|=4$}\end{tabular}} & \cellcolor[HTML]{EFEFEF}\begin{tabular}[c]{@{}c@{}}\textbf{$p=7.8\cdot10^{-02}$}\\  \textbf{$p_{hw}=0.612$}\\  \textbf{$|x_{out}|=4$}\end{tabular} \\ \hline\hline
                                                                                               & \multicolumn{1}{c|}{base}                  & DEHB                  & \multicolumn{1}{c|}{\begin{tabular}[c]{@{}c@{}}$p=4.0\cdot10{-243}$\\  $p_{hw}=0.004$\\  $|x_{out}|=5$\end{tabular}} & \multicolumn{1}{c|}{\begin{tabular}[c]{@{}c@{}}$p=4.0\cdot10^{-80}$\\  $p_{hw}=0.937$ \\ $|x_{out}|=9$\end{tabular}}                                                    & \multicolumn{1}{c|}{\begin{tabular}[c]{@{}c@{}}$p=1.8\cdot10^{-08}$\\  $p_{hw}=0.079$\\  $|x_{out}|=6$\end{tabular}}                                                    & \begin{tabular}[c]{@{}c@{}}$p=3.4\cdot10^{-03}$\\  $p_{hw}=0.191$\\  $|x_{out}|=6$\end{tabular}                                                    \\ \cline{2-7} 
                                                                                               & \multicolumn{1}{c|}{DEHB}                  & base                  & \multicolumn{1}{c|}{\textbf{best}}                                                                                   & \multicolumn{1}{c|}{\begin{tabular}[c]{@{}c@{}}$p=9.3\cdot10^{-03}$\\  $p_{hw}=0.056$\\  $|x_{out}|=8$\end{tabular}}                                                    & \multicolumn{1}{c|}{\textbf{best}}                                                                                                                                      & \textbf{best}                                                                                                                                      \\ \cline{2-7} 
                                                                                               & \multicolumn{1}{c|}{DEHB}                  & RS                    & \multicolumn{1}{c|}{\begin{tabular}[c]{@{}c@{}}$p=6.8\cdot10^{-04}$\\  $p_{hw}=0.051$\\  $|x_{out}|=6$\end{tabular}} & \multicolumn{1}{c|}{-}                                                                                                                                                  & \multicolumn{1}{c|}{\cellcolor[HTML]{EFEFEF}\begin{tabular}[c]{@{}c@{}}\textbf{$p=7.7\cdot10^{-02}$}\\  \textbf{$p_{hw}=0.109$}\\  \textbf{$|x_{out}|=5$}\end{tabular}} & -                                                                                                                                                  \\ \cline{2-7} 
\multirow{-4}[20]{*}{\begin{tabular}[c]{@{}c@{}}Google Brax \\ Ant\\ (maximise)\end{tabular}}      & \multicolumn{2}{c|}{DEHB (ours)}                                   & \multicolumn{1}{c|}{\begin{tabular}[c]{@{}c@{}}$p=1.7\cdot10^{-02}$\\  $p_{hw}=0.894$\\  $|x_{out}|=2$\end{tabular}} & \multicolumn{1}{c|}{\textbf{best}}                                                                                                                                      & \multicolumn{1}{c|}{\cellcolor[HTML]{EFEFEF}\begin{tabular}[c]{@{}c@{}}\textbf{$p=7.7\cdot10^{-01}$}\\  \textbf{$p_{hw}=0.978$}\\  \textbf{$|x_{out}|=6$}\end{tabular}} & \cellcolor[HTML]{EFEFEF}\begin{tabular}[c]{@{}c@{}}\textbf{$p=6.4\cdot10^{-01}$}\\  \textbf{$p_{hw}=0.170$}\\  \textbf{$|x_{out}|=5$}\end{tabular} \\ \hline\hline
                                                                                               & \multicolumn{1}{c|}{base}                  & DEHB                  & \multicolumn{1}{c|}{\begin{tabular}[c]{@{}c@{}}$p=1.1\cdot10^{-27}$\\  $p_{hw}=0.123$\\  $|x_{out}|=7$\end{tabular}} & \multicolumn{1}{c|}{\begin{tabular}[c]{@{}c@{}}$p=2.2\cdot10^{-26}$\\  $p_{hw}=0.797$\\  $|x_{out}|=6$\end{tabular}}                                                    & \multicolumn{1}{c|}{\begin{tabular}[c]{@{}c@{}}$p=1.0\cdot10^{-04}$\\  $p_{hw}=0.293$\\  $|x_{out}|=6$\end{tabular}}                                                    & \begin{tabular}[c]{@{}c@{}}$p=3.5\cdot10^{-06}$\\  $p_{hw}=0.773$\\  $|x_{out}|=4$\end{tabular}                                                    \\ \cline{2-7} 
                                                                                               & \multicolumn{1}{c|}{DEHB}                  & base                  & \multicolumn{1}{c|}{\begin{tabular}[c]{@{}c@{}}$p=3.8\cdot10^{-10}$\\  $p_{hw}=0.086$\\  $|x_{out}|=7$\end{tabular}} & \multicolumn{1}{c|}{\begin{tabular}[c]{@{}c@{}}$p=4.8\cdot10^{-23}$\\  $p_{hw}=0.780$\\  $|x_{out}|=8$\end{tabular}}                                                    & \multicolumn{1}{c|}{\begin{tabular}[c]{@{}c@{}}$p=1.6\cdot10^{-14}$\\  $p_{hw}=0.025$\\  $|x_{out}|=7$\end{tabular}}                                                    & \begin{tabular}[c]{@{}c@{}}$p=3.1\cdot10^{-51}$\\  $p_{hw}=0.018$\\  $|x_{out}|=6$\end{tabular}                                                    \\ \cline{2-7} 
                                                                                               & \multicolumn{1}{c|}{DEHB}                  & RS                    & \multicolumn{1}{c|}{\begin{tabular}[c]{@{}c@{}}$p=1.5\cdot10^{-02}$\\  $p_{hw}=0.094$\\  $|x_{out}|=4$\end{tabular}} & \multicolumn{1}{c|}{-}                                                                                                                                                  & \multicolumn{1}{c|}{\cellcolor[HTML]{EFEFEF}\begin{tabular}[c]{@{}c@{}}\textbf{$p=9.9\cdot10^{-01}$}\\  \textbf{$p_{hw}=0.132$}\\  \textbf{$|x_{out}|=5$}\end{tabular}} & -                                                                                                                                                  \\ \cline{2-7} 
\multirow{-4}[20]{*}{\begin{tabular}[c]{@{}c@{}}Google Brax \\ Humanoid\\ (maximise)\end{tabular}} & \multicolumn{2}{c|}{DEHB (ours)}                                   & \multicolumn{1}{c|}{\textbf{best}}                                                                                   & \multicolumn{1}{c|}{\textbf{best}}                                                                                                                                      & \multicolumn{1}{c|}{\textbf{best}}                                                                                                                                      & \textbf{best}                                                                                                                                      \\ \hline\hline
                                                                                                   & \multicolumn{1}{c|}{base}                  & DEHB                  & \multicolumn{2}{c|}{}                                                                                                                                                                                                                                                                      & \multicolumn{1}{c|}{\begin{tabular}[c]{@{}c@{}}$p=2.1\cdot10^{-45}$\\  $p_{hw}=0.001$\\  $|x_{out}|=3$\end{tabular}}                                                    & \begin{tabular}[c]{@{}c@{}}$p=3.5\cdot10^{-82}$\\  $p_{hw}=0.003$\\  $|x_{out}|=5$\end{tabular}                                                                                                                                                  \\ \cline{2-3} \cline{6-7}
                                                                                               & \multicolumn{1}{c|}{DEHB}                  & base                  & \multicolumn{2}{c|}{}                                                                                                                                                                                                                                                                          & \multicolumn{1}{c|}{\begin{tabular}[c]{@{}c@{}}$p=4.8\cdot10^{-13}$\\  $p_{hw}=0.278$\\  $|x_{out}|=3$\end{tabular}}                                                    & \textbf{best}                                                                                                                                               \\ \cline{2-3} \cline{6-7}
                                                                                               & \multicolumn{1}{c|}{DEHB}                  & RS                    & \multicolumn{2}{c|}{}                                                                                                                                                                                                                                                                          & \multicolumn{1}{c|}{\begin{tabular}[c]{@{}c@{}}$p=4.4\cdot10^{-06}$\\  $p_{hw}=0.022$\\  $|x_{out}|=5$\end{tabular}}                                                    & -                                                                                                                                                  \\ \cline{2-3} \cline{6-7}
\multirow{-4}[20]{*}{\begin{tabular}[c]{@{}c@{}}Robosuite \\ Wipe\\ (maximise)\end{tabular}}       & \multicolumn{2}{c|}{DEHB (ours)}                                   & \multicolumn{2}{c|}{\multirow{-4}[20]{*}{-}}                                                                                                                                                                                                                                               & \multicolumn{1}{c|}{\textbf{best}}                                                                                                                                      & \cellcolor[HTML]{EFEFEF}\begin{tabular}[c]{@{}c@{}}$p=3\cdot10^{-1}$\\  $p_{hw}=0.296$\\  $|x_{out}|=4$\end{tabular}                                                                                                                                                                                                           \\ \hline
\end{tabular}

    \caption{We show the p-values for significance in the difference of each experiment to the best-performing optimisation corresponding to the results presented in Table \ref{tab:external_objective}.
    Additionally, $p_{wh}$ gives the p-values of White's test for heteroskedasticity and $|x_{out}|$ the number of removed outliers from the 100 data points for each comparison.
    We require homoskedasticity of the residuals and hence values of $p_{wh} > 0.05$.
    Highlighted cells do not perform significantly differently from the best corresponding optimisation experiment.
    }
    \label{tab:lme_stats}
\end{table}
The values correspond to comparisons against the cell identified as the best performer. 
Entries that do not exhibit a significant difference from the best-performing cells are highlighted. 
The assumption of homoskedasticity is satisfied in the majority of cases, as indicated by p-values greater than $0.05$. 
This is particularly true in scenarios where the performance of the experiment is not significantly different from that of the best optimisation.
Moreover, the p-values from the Wald test are generally well above the significance threshold of 0.05 when there is no significant difference. 
In our comparisons, the normality assumption was met in 18 out of 35 cases. 
Given the mixed-effect models' resilience to deviations from the normality assumption and the large number of cases where this assumption was satisfied, we conclude that our test results are reliable for comparing the outcomes of our experiments.

\end{document}